\documentclass[10pt,twocolumn,letterpaper]{article}

\usepackage{cvpr}              

\usepackage{multirow}
\usepackage{colortbl}
\usepackage{booktabs}
\usepackage{fontawesome5}
\usepackage{makecell}
\usepackage{graphicx}
\usepackage[accsupp]{axessibility} 

\usepackage[dvipsnames]{xcolor}

\definecolor{cvprblue}{rgb}{0.21,0.49,0.74}
\usepackage[pagebackref,breaklinks,colorlinks,citecolor=cvprblue]{hyperref}

\def\paperID{1203} 
\def\confName{CVPR}
\def\confYear{2024}

\title{BigGait: Learning Gait Representation You Want by Large Vision Models}

\author{
Dingqiang Ye$^{1,2*}$, 
Chao Fan$^{1,2}$\thanks{Equal contribution.}, 
Jingzhe Ma$^{1,2}$, 
Xiaoming Liu$^3$, 
and Shiqi Yu$^{1,2}$\thanks{Corresponding author.} \\
{\normalsize $^1$ Research Institute of Trustworthy Autonomous System, Southern University of Science and Technology, Shenzhen, China}\\
{\normalsize $^2$ Department of Computer Science and Engineering, Southern University of Science and Technology, Shenzhen, China}\\
{\normalsize $^3$ Michigan State University, Michigan, United States} \\
{\tt\scriptsize \{11810121, 12131100, 12031127\}@mail.sustech.edu.cn, liuxm@cse.msu.edu, yusq@sustech.edu.cn
}
}

\begin{document}
\maketitle
\begin{abstract}
Gait recognition stands as one of the most pivotal remote identification technologies and progressively expands across research and industry communities.
However, existing gait recognition methods heavily rely on task-specific upstream driven by supervised learning to provide explicit gait representations like silhouette sequences, which inevitably introduce expensive annotation costs and potential error accumulation.
Escaping from this trend, this work explores effective gait representations based on the all-purpose knowledge produced by task-agnostic Large Vision Models (LVMs) and proposes a simple yet efficient gait framework, termed \textbf{BigGait}.
Specifically, the Gait Representation Extractor (GRE) within BigGait draws upon design principles from established gait representations, effectively transforming all-purpose knowledge into implicit gait representations without requiring third-party supervision signals.
Experiments on CCPG, CAISA-B* and SUSTech1K indicate that BigGait significantly outperforms the previous methods in both within-domain and cross-domain tasks in most cases, and provides a more practical paradigm for learning the next-generation gait representation.
Finally, we delve into prospective challenges and promising directions in LVMs-based gait recognition, aiming to inspire future work in this emerging topic.
The source code is available at \url{https://github.com/ShiqiYu/OpenGait}. 
\end{abstract}    
\section{Introduction}
\label{sec:intro}

Vision-based gait recognition aims to identify individuals based on their unique walking patterns. 
Compared to other biometric modalities like face, fingerprint, and iris, gait stands out for its non-intrusive nature and the ability to identify individuals at long distances without the need for active cooperation.
These distinctive advantages make gait recognition exceptionally suitable for a range of security applications, like suspect tracking and identity investigation~\cite{nixon2006automatic}.

\begin{figure}[t]
\centering
\includegraphics[height=6.5cm]{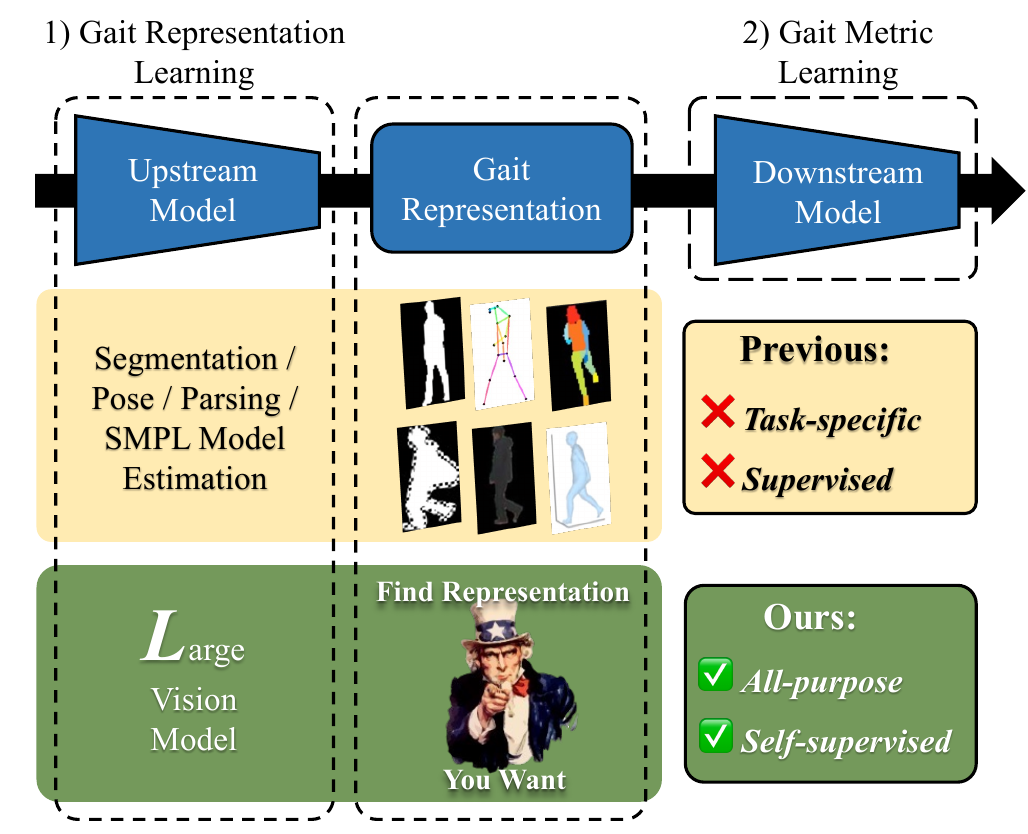}
\caption{The upstream and downstream parts of existing gait recognition methods are responsible for gait representation and metric learning, respectively. }
\label{fig:intro}
\vspace{-1em}
\end{figure}

Existing gait recognition methods~\cite{Chao2019, liao2017pose, zheng2023parsing, liang2022gaitedge, li2020end, zhang2019gait, 9154576, FarSight2024} heavily depend on upstream tasks driven by supervised learning, as illustrated by Fig.~\ref{fig:intro}. 
These tasks encompass a range of objectives, including, but not limited to, pedestrian segmentation~\cite{wang2020deep}, human parsing~\cite{liu2022cdgnet}, body posture~\cite{cao2017realtime}, and SMPL~\cite{loper2023smpl} model estimation. 
In general, the upstream model serves the purpose of providing task-specific priors to filter out the gait-irrelevant cues within walking videos, especially for RGB-encoded background and texture characteristics.
Next, some inductive biases may be utilized to refine the intermediate gait representation, ranging from basic operations like size alignment~\cite{Takemura2018}, coordinate normalization~\cite{fu2023gpgait} to sophisticated ones like making the silhouette edges differentiable~\cite{liang2022gaitedge} and enforcing the appearance reconstruction~\cite{zhang2019gait, li2020end}.
Briefly, existing gait representations are largely predetermined by upstream supervised tasks. 

Escaping from the reliance on upstream task-specific models, this paper makes a pioneering effort to acquire desired gait features from \textit{task-agnostic Large Vision Models} (LVMs).
We are strongly driven by the following insights:
\begin{itemize}
    \item The recent breakthroughs~\cite{oquab2023dinov2, kirillov2023segment, radford2021learning, goyal2021self} have confirmed the discriminability and generalization of all-purpose features produced by LVMs, showcasing an attractive opportunity to enhance gait metric learning. 
    \item Self-supervised pre-training of LVMs obviates the need for labeled datasets required for training the state-of-the-art gait upstream models, thus eliminating the substantial costs necessary for annotating elements such as the silhouette, skeleton, and more, on a large scale.
    \item The task-agnostic knowledge embedded in LVMs is learned from web-scale datasets  without task-specific supervision, thus ideally avoiding error accumulation imposed by specific upstream tasks to a large extent. 
\end{itemize}

\begin{figure}[t]
\centering
\includegraphics[width=0.45\textwidth]{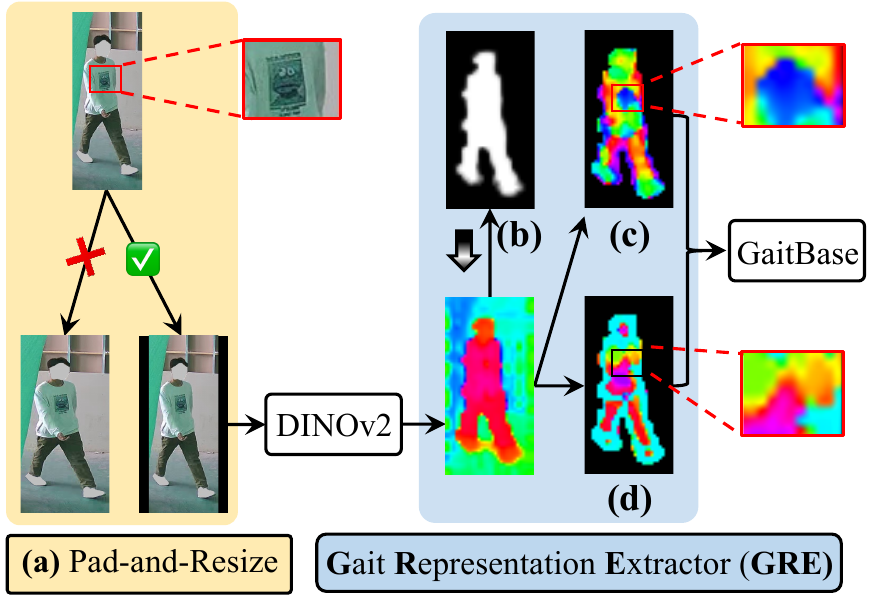}
\caption{The illustration of (a) body proportion preservation trick is provided in the Supplementary Material. 
(b)-(d) respectively present the visualization of intermediate representation generated by (b) mask, (c) appearance, and (d) denoising branch.}
\label{fig:fig2}
\vspace{-5mm}
\end{figure}

To bring the above visions to life, we propose the first LVMs-based gait framework termed \textit{BigGait}\footnote{Both `large' and `big' can refer to LVMs' size. We use the latter since it is relatively common in daily conversations.}.
Structurally, we configure the upstream and downstream models as DINOv2~\cite{oquab2023dinov2} and GaitBase~\cite{fan2022opengait} with minor adjustments, thanks to their representativeness in the domains of LVMs and gait research, respectively. 
In this paper, we regard LVMs-based gait recognition as a challenging task focusing on transforming all-purpose features into effective gait representations. 
To achieve this goal, we propose the \textit{Gait Representation Extractor} (GRE) bridging the upstream and downstream models, as illustrated in Fig.~\ref{fig:fig2}:
\begin{itemize}
    \item \textbf{Mask Branch}. Borrowing the design of the silhouette, as shown in Fig.~\ref{fig:fig2} (b), GRE develops a mask branch to autonomously infer the foreground without supervision, thus excluding the background interference on the whole. 
    \item \textbf{Appearance Branch}. As observed in Fig.~\ref{fig:fig2} (c), background removal makes foreground features diverse and discriminative over body parts, resulting in a parsing-like representation. However, this branch may potentially introduce texture-related noise. 
    \item \textbf{Denoising Branch} in Fig.~\ref{fig:fig2} (d) introduces smoothness constraints along the spatial dimension to reduce the high-frequency textural characteristics. Moreover, a diversity constraint is introduced to prevent trivial solutions.
\end{itemize}

Overall, our GRE module adopts the design principles of some established gait representations and distinguishes itself by leveraging all-purpose knowledge and removing upstream task-specific supervision.
Technically, GRE relies solely on soft constraints to extract effective gait representations, \textit{i.e.}, capturing truthful body structural characteristics while effectively excluding gait-irrelevant noise. 
Moreover, this work also finds that the utilization of soft constraints may introduce two major challenges:
(1) \textbf{Interpretability}. 
In contrast to the more visually intuitive modality such as the silhouette and skeleton, which directly convey physical meanings, the learned representation derived from soft geometrical constraints, presented as the multi-channel feature map, may not offer the same level of intuitiveness.
(2) \textbf{Purity}. 
Completely suppressing the texture characteristics of RGB videos through soft constraints remains a challenge for gait recognition~\cite{liang2022gaitedge}. 
Our comprehensive cross-clothing and cross-domain evaluations provide strong evidence of the dominant role of gait features within the acquired gait representations.
Nevertheless, we find that data distribution may still influence outcomes.

In summary, this paper presents a pioneering attempt to open up the promising LVMs-based gait recognition research. 
The main contributions can be outlined as follows: 
\begin{itemize}
    \item BigGait charts a groundbreaking and practical learning paradigm for the next-generation gait representation, with gait guidance shifting from task-specific priors to LVMs-based all-purpose knowledge.
    \item We establish the effectiveness of all-purpose knowledge for gait description, as evidenced by its discriminability and generalization across various gait datasets, including CCPG~\cite{li2023depth}, CASIA-B*~\cite{yu2006framework} and SUSTech1K~\cite{Shen_2023_CVPR}.
    BigGait achieves the SoTA performance in most cases. 
\end{itemize}
\begin{figure*}[t]
\centering
    \includegraphics[height=5.8cm]{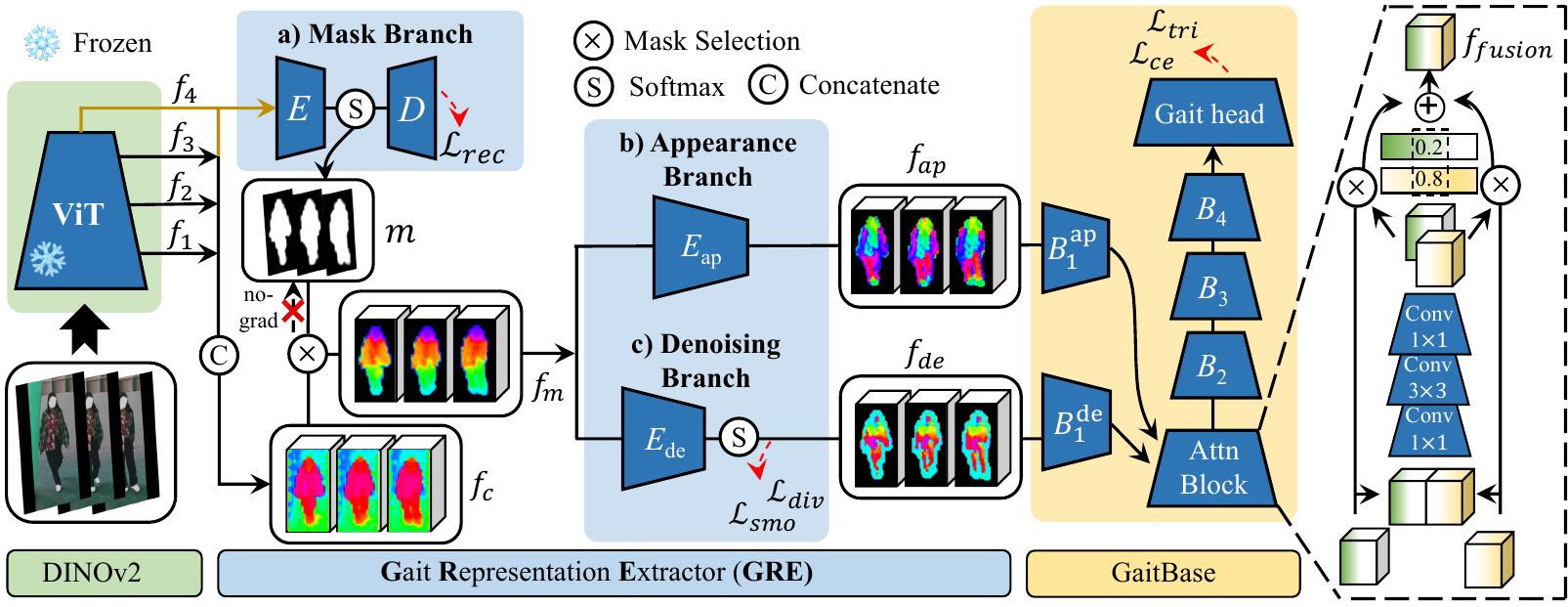}
\caption{The workflow of BigGait. Specifically, the upstream model is instantiated as DINOv2 aiming to produce all-purpose features. The central gait representation extractor (GRE) owns three branches respectively responsible for the background removal, feature transformation, and feature refining. In the end, the modified GaitBase is employed for gait metric learning.
}
\vspace{-1.5em}
\label{fig:overview}
\end{figure*}

\section{Related Works}
\label{sec:related_works}

\noindent \textbf{Upstream Models for Gait Representation Learning.}
Walking videos often contain numerous extraneous elements, including background distractions and variations in color, texture, clothing, and carried items within the foreground.
Prior methods have introduced a range of task-specific models to address these challenges. These models are designed to provide binary silhouettes~\cite{Chao2019, fan2020gaitpart, lin2021gait, fan2022opengait}, body skeletons~\cite{liao2017pose, teepe2021gaitgraph}, human parsing~\cite{zheng2023parsing}, SMPL models~\cite{li2020end}, or a combination of these representations~\cite{zheng2022gait3d}, to serve as the input for downstream gait models.
Moreover, several end-to-end methods~\cite{zhang2019gait, liang2022gaitedge} aim to optimize the intermediate gait representation globally. However, it is important to recognize that these upstream tasks heavily rely on supervised learning, giving rise to three primary challenges: a) considerable annotation costs, b) restricted form of gait representation, and c) potential accumulation of estimation errors.
In this paper, BigGait instantiates the upstream model as a self-supervised LVM, constructing the desired gait representation by leveraging the all-purpose knowledge in a novel manner.

\noindent \textbf{Downstream Models for Gait Metric Learning.}
Depending on the gait representations used, downstream gait models are designed accordingly.
For instance, Convolutional Neural Networks (CNNs) are extensively employed for modeling image-based gait representations such as silhouettes and human parsing, while Graph Convolutional Networks (GCNs) are commonly employed for learning skeleton-based representations.
In the majority of cases, downstream gait models place their emphasis on local and global temporal modeling~\cite{huang2021context, 3DLocal, dou2022metagait}, along with the creation of spatially hierarchical descriptions~\cite{fan2020gaitpart, lin2021gait, fu2023gpgait}.

\noindent \textbf{Large Vision Models.}
Inspired by the success of LLMs~\cite{devlin2018bert, brown2020language}, researchers have ventured into exploring foundation models in the field of computer vision~\cite{oquab2023dinov2, kirillov2023segment, radford2021learning}, \textit{i.e.}, learning generalizable features from web-scale data sources by scalable vision models. 
For example, 
CLIP~\cite{radford2021learning} used a form of textual supervision to guide the training of visual representation. 
SAM~\cite{kirillov2023segment} developed a promotable model and pre-trained it on a broad dataset using a task that enables powerful generalization for image segmentation. 
This paper focuses on another alternative termed self-supervised LVMs~\cite{chen2020simple, simsiam, he2022masked}, where features are learned from images alone without supervision. 
Specifically, the employed DINOv2~\cite{oquab2023dinov2} demonstrated that self-supervised learning has the potential to learn all-purposed visual features if pre-trained on a large quantity of curated data. 
Here, all-purpose features aim to work out of the box on any task, \textit{e.g.}, image classification, and dense recognition like semantic segmentation and depth estimation. 
The main goal of this paper is to learn gait representation from these all-purpose features, thereby benefiting from the advantages of LVMs. 

\noindent \textbf{Other Related Works.}
Recently, GaitSSB~\cite{fan2023learning} introduced a self-supervised pre-training benchmark tailored for silhouette-based gait recognition. Notably, this pre-training approach is gait-specific and directly influences the downstream model. It still heavily relies on an upstream supervised segmentation model, whereas in the case of DINOv2 within our BigGait, the pre-training is task-agnostic and serves as the upstream model without the need for supervision. Another distinctive advantage of BigGait lies in the multi-branch architecture of the GRE module, which offers a gait representation resembling the capabilities of multi-modality, even though it derives from a single source.
While typical multi-modal methods~\cite{zheng2022gait3d, peng2023learning} require multiple supervised upstream models, we can create diverse gait representations or their combinations relying solely on all-purpose features obtained from self-supervised LVMs. 

\section{Method}
\label{sec:method}
As shown in Fig.~\ref{fig:overview}, BigGait consists of three parts, \textit{i.e.}, the upstream frozen DINOv2~\cite{oquab2023dinov2}, the central GRE module, and the downstream adjusted GaitBase~\cite{fan2022opengait}, respectively responsible for all-purpose feature extraction, gait representation construction, and gait metric learning. 

\subsection{Overview}
\label{sec:method_overview}
BigGait takes an RGB video as input and processes each frame in parallel. For simplicity, we focus on a single frame in the following description.
Before feeding the image into the upstream model, this work adopts a Pad-and-Resize trick to resize it into a fixed resolution of $448\times224$ while keeping the truthful body proportions.
The illustration of this detail is provided in the Supplementary Material.


\noindent \textbf{Upstream Model.}
The upstream DINOv2 owns a scalable ViT~\cite{dosovitskiy2020image} as the backbone, with parameter counts ranging from $21$M to $1,100$M. In this study, we opt for the smallest and second largest counterparts, namely, ViT-S/14 ($21$M) and ViT-L/14 ($302$M), to construct BigGait-S and BigGait-L, respectively. 
Given the resized RGB image, the ViT first divides it into non-overlapping patches with a size of $14\times14$, thus taking $32\times16$ tokenized vectors as input. 
Next, the positional encodings are added, and subsequent stacked ViT blocks transform the features. 
We gather tokens from low to high layers with uniform intervals, and all these $32\times16$ tokens are concatenated according to the spatial correspondence to form a feature map. 
These details follow the official implementation~\cite{oquab2023dinov2}. 
Finally, we upsample the feature map to $64\times32$ as output.

We leverage the officially provided model checkpoint pre-trained on LVD-142M~\cite{oquab2023dinov2} and freeze the imported parameters during gait metric learning.
To reduce carbon emissions, we establish BigGait-S as the reference version, even though BigGait-L exhibits slightly higher performance. 
Thanks to this choice, the upstream part of BigGait is even smaller than many typical segmentation, human parsing, and pose estimation networks widely used by existing gait methods, as shown in Tab.~\ref{tab: parameter size}.

\begin{table}[t!]
\centering
\caption{Parameter and GFLOPs of different upstream/overall methods. The GFLOPs are calculated for the input size $448\times224$.}
\vspace{-2.mm}
\renewcommand{\arraystretch}{0.9}
\resizebox{0.99\columnwidth}{!}{

\begin{tabular}{ccccc} 
\toprule[2pt]
Upstream Obj.                                         & Methods                                                         & Backbone  & \#Params (M) & GFLOPs  \\ 
\midrule[1pt]
\multirow{2}{*}{Segmentation}                & UNet3+~\cite{huang2020unet}                    & UNet3+    & $27.0$    & $308.2$   \\
                                             & DeepLabV3+~\cite{chen2018encoder}              & ResNet-50 & $26.8$    & $43.7$    \\
\multirow{2}{*}{Human Parsing}               & SCHP~\cite{li2020self}                         & A-CE2P    & $66.6$    & $33.5$    \\
                                             & HRNet~\cite{wang2020deep}                      & HRNet-W48 & $70.1$    & $61.9$    \\
\multirow{2}{*}{Pose Estimation}             & HRNet~\cite{sun2019deep}                       & HRNet-W32 & $28.5$    & $15.8$    \\
                                             & DEKR~\cite{geng2021bottom}                     & HRNet-W32 & $29.5$    & $17.1$    \\ 
\multirow{2}{*}{All-purpose (\textbf{Ours})} & \multirow{2}{*}{DINOv2~\cite{oquab2023dinov2}} & ViT-S/14  & $21.5$    & $11.0$    \\
                                             &                                                                 & ViT-L/14  & $302.9$   & $155.3$  \\
\toprule


\toprule
Overall                & Upstream                & Downstream & \#Params (M) & GFLOPs  \\ 
\midrule
Sil.-based       & DeepLabV3+  & GaitBase   & $34.2$    & $45.4$    \\
Parsing-based          & SCHP           & GaitBase   & $74.0$    & $35.2$    \\
Skeleton-based         & HRNet-W32     & Gait-TR    & $29.0$      & $31.2$    \\
BigGait~(\textbf{Ours}) & ViT-S/14 +~GRE & GaitBase   & $30.8$    & $12.7$    \\
\bottomrule[2pt]
\end{tabular}
}
\vspace{-6mm}
\label{tab: parameter size}
\end{table}

\noindent \textbf{Central Module.}
Directly using the learned all-purpose features can result in inferior performances. 
In light of this, this paper pays primary attention to transforming all-purpose features into useful gait representations, thereby proposing the GRE module. 
Specifically, the GRE module includes three primary components, \textit{i.e.}, the mask, appearance, and denoising branches respectively responsible for the background removal, feature transformation, and feature refining.
This module outputs two types of gait representations, and Sec.~\ref{sec:method_GRE} will provide details. 

\noindent \textbf{Downstream Model.}
We make a minor modification to GaitBase~\cite{fan2022opengait} so that it can process the two-stream input.
As shown in Fig.~\ref{fig:overview}, $B_1, B_2, B_3$ and $B_4$ refer to the standard blocks within GaitBase, where $B_1^{\text{ap}}$ and $B_1^{\text{de}}$ own the different parameters but share the identical structure with $B_1$. 
Then, an attention block possessing the widely used cross-and-select architecture~\cite{li2019selective} targets aggregating the generated two-branch features. 
Excluding the above modifications, our implementation of GaitBase makes no architectural difference from the official one~\cite{fan2022opengait}.
The training is driven by the triplet $L_{tri}$ and cross-entropy $L_{L_{ce}}$ losses. 

\noindent \textbf{Visualization.} 
To understand the content of each intermediate feature map, namely $f_c, f_m, f_{ap}$, and $f_{de}$ depicted in Fig.~\ref{fig:overview}, we conduct a PCA using the pixel-wise features from the entire dataset.
Projecting pixel-wise features onto the first three PCA bases allow us to visualize a high-dim feature map as an RGB color image.
This visualization approach is inherited from DINOv2~\cite{oquab2023dinov2}.

\subsection{Gait Representation Extractor}
\label{sec:method_GRE}
As shown in Fig.~\ref{fig:overview}, 
$f_1, f_2, f_3, f_4$ denote the feature map generated by various stages of the ViT backbone with the corresponding semantic hierarchy spanning from low to high levels, which is a common practice suggested by DINOv2. 
Each of them is $2\times$ upsampled by bilinear interpolation to improve the resolution, \textit{i.e.}, exhibited as a 3-D tensor with a size of $384\times64\times32$ while the first dimension denotes the output channel of the upstream DINOv2. 
Additionally, we concatenate these feature maps along the channel dimension to form $f_c$, \textit{i.e.}, with a size of $1,536\times64\times32$, which is rich in both low-level details and high-level semantics. 
As shown in Fig.~\ref{fig:overview}, $f_c$ is dominated by the foreground and background regions accompanied by noisy spots.

\noindent \textbf{Mask Branch.}
Inspired by the design of the silhouette, the GRE module develops a mask branch to remove the background noise on the whole. 
Specifically, an auto-encoder is proposed to generate the foreground mask based on $f_4$: 
\vspace{-2.mm}
\begin{equation}
    \begin{aligned}
    m &= \text{softmax}(E(f_4)) \\ 
    \bar{f}_4 &= D(m) \\ 
    L_{rec} &= \left\| f_4 - \bar{f}_4 \right\|_2 ,
    \end{aligned}
\vspace{-2.mm}
\end{equation}
where $E$ and $D$ denote the linear convolution layer with a kernel size of $1\times1$ and the output channel of $2$ and $384$, respectively. 
The softmax function is conducted along the channel dimension, and $L_{rec}$ presents the reconstruction loss. 
Here, we use $f_4$ rather than $f_c$ due to its better empirical results.
The cause may be that $f_4$ contains higher-level semantically separable features.

\begin{figure}[t]
\centering
    \includegraphics[height=0.18\columnwidth]{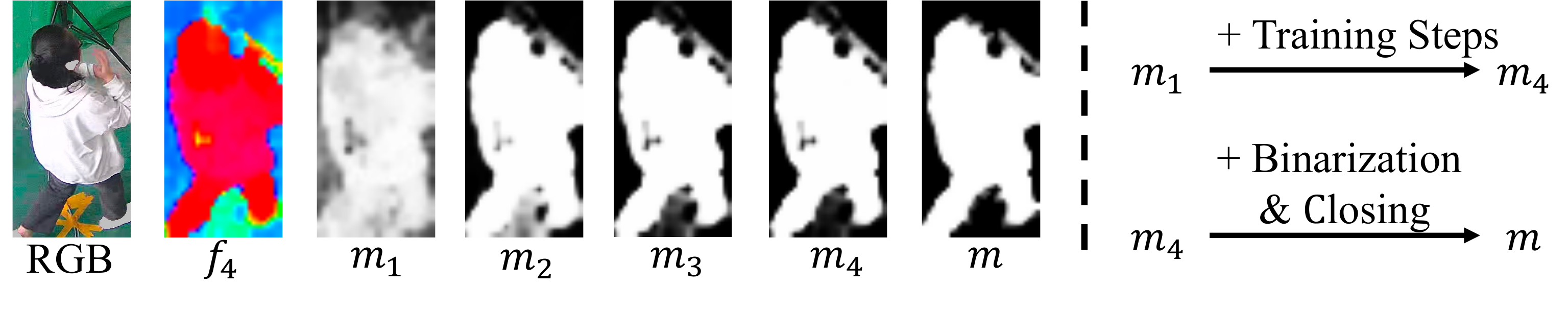}
\vspace{-4mm}
\caption{Visualization of unsupervised mask learning.}
\label{fig: mask_branch}
\vspace{-2em}
\end{figure}

Intuitively, the mask branch works like a PCA by searching two mutually exclusive first components. 
Thanks to the segmentable nature of $f_4$, as shown in Fig.~\ref{fig: mask_branch}, the mask $m$ can be easily learned in this way. 
Notably, the $m$ is a two-channel tensor, and which channel represents the foreground is uncertain. 
So we select the channel whose activations lay more in the image center as the foreground mask. 
Moreover, we perform the binarization and closing operations to reduce the potential cavities and breakpoints. 
These common tricks do not appear in Fig.~\ref{fig:overview} for brevity, \textit{i.e.}, the symbol of $m$ in Fig.~\ref{fig:overview} denotes the final foreground mask. 

After masking the background regions in $f_c$ with $m$, we obtain $f_m$: 
\vspace{-2mm}
\begin{equation}
    \begin{aligned}
    f_{m} = m \cdot f_{c} ,
    \end{aligned}
\vspace{-2.mm}
\end{equation}
where $\cdot$ denote the multiplication.
Fig.~\ref{fig:overview} shows that this process not only largely filters out the background noise but also makes foreground features diverse and discriminative over body parts, resulting in a parsing-like effect. 

\noindent \textbf{Appearance Branch.}
To effectively capture characteristics within $f_m$, GRE develops an appearance branch: 
\vspace{-2.mm}
\begin{equation}
    f_{ap} = E_{\text{ap}}(f_m) ,
\vspace{-2.mm}
\end{equation}
where $E_{\text{ap}}$ denotes a linear convolution layer with the kernel size of $1\times1$ and output channel of $C$.
Here $C$ is a hyper-parameter determining the channel of intermediate gait representation $f_{ap}$.
Note the functionality of $E_{\text{ap}}$ can be easily approximated by the early layers of GaitBase. 
Here we introduce this layer mainly for GPU memory saving by ensuring $C\ll1,536$ (the channel number of $f_m$).
Overall, $E_{\text{ap}}$ linearly adjusts features along the channel dimension and then directly feeds its output $f_{ap}$ into the gait model. 


By comparing the input $f_m$ with the output $f_{ap}$, as shown in Fig.~\ref{fig:overview}, the emergence of noisy elements within the latter should be primarily driven by the downstream gait model rather than the linear $E_{\text{ap}}$. 
This observation is consistent with findings in related studies~\cite{liang2022gaitedge, li2023depth}, suggesting that the gait model tends to emphasize the clearly unchanged gait-irrelevant cues rather than the subtle gait patterns. 
Further visualizations in Sec.~\ref{Visualization Analysis} reveal the correlation between these noises and the clothing texture on the body.

To alleviate this issue, the GRE module develops a denoising branch parallel to this appearance branch. 

\noindent \textbf{Denoising Branch.}
This branch regards texture noises as high-frequency signals along spatial dimensions, thus introducing a smoothness loss $L_{smo}$, formulated as: 
\vspace{-2.mm}
\begin{equation}
    \begin{aligned}
        f_{de} &= \text{softmax}(E_{\text{de}}(f_m)) \\
        L_{smo} &= | \text{sobel}_\text{x} * f_{de} | + | \text{sobel}_\text{y} * f_{de}|  ,
    \end{aligned}
    \label{equ:smo}
\vspace{-2.mm}
\end{equation}
where $E_{\text{de}}$ denotes non-linear layers composed by a $1\times1$ convolution, a batch normalization, a GELU, and an additional $1\times1$ convolution. 
Its output channel is set to $C$ which is the same as that of the appearance branch for brevity. 
The softmax function is performed along the channel dimension to normalize the features. 
The smoothness loss is defined by the absolute value of channel-wise image gradients along both the width and height axes. 
In Eq.~\ref{equ:smo}, $\text{sobel}_x$ and $\text{sobel}_y$ refer to the x-axis and y-axis Sobel operator~\cite{sobel19683x3}, and the symbol of $*$ represents the convolution operation. 

From a definitional standpoint, $L_{smo}$ can not remove the gradual textural noise effectively. 
But for gait recognition, we consider that the evident textures on body images, \textit{e.g.}, the clothing logo and pattern, are generally high-frequency characteristics that can be easily eliminated by $L_{smo}$.

Relying solely on the smoothness loss $L_{smo}$ can potentially lead to a trivial solution, where each pixel of $f_{de}$ essentially presents nearly identical feature vectors, resulting in a significant loss of representational diversity. To tackle this issue, we introduce an additional diversity loss based on information entropy.
The softmax normalization ensures a certain probability distribution of activations across the channel dimension, and the uniform one corresponds to the maximum information entropy. 
Thus, the diversity loss is: 
\vspace{-3.mm}
\begin{equation}
    \begin{aligned}
        p_i &= \text{sum}(f_{de}^i) / \sum_{i=1}^{C} \text{sum}(f_{de}^i) \\
        L_{div} &= \text{log}C + \sum_{i=1}^{C} p_i\text{log}p_i ,
    \end{aligned}
\vspace{-3.mm}
\end{equation}
where $f_{de}^i$ and $p_i$ respectively represents the activation map of the $i$-th channel and the corresponding frequency to entire activations. 
The constant term $\text{log}C$ denotes the maximum entropy and is imported to prevent the negative loss. 

Overall, the learning of the denoising branch is entirely driven by the above two soft constraints. 
These losses only perform in the foreground produced by the mask branch.
Finally, we fuse $f_{ap}$ and $f_{de}$ using attention weights:
\vspace{-2mm}
\begin{equation}
    \begin{aligned}
        f_{fusion} = Attn(\text{B}_1^{ap}(f_{ap}), \text{B}_1^{de}(f_{de})),
    \end{aligned}
\vspace{-2.mm}
\end{equation}
where $Attn$ is an attention block, following~\cite{fan2023skeletongait}.

\subsection{Visualization of Intermediate Representation}
\label{Visualization Analysis}
\begin{figure}[t]
\centering
    \includegraphics[width=0.85\columnwidth]{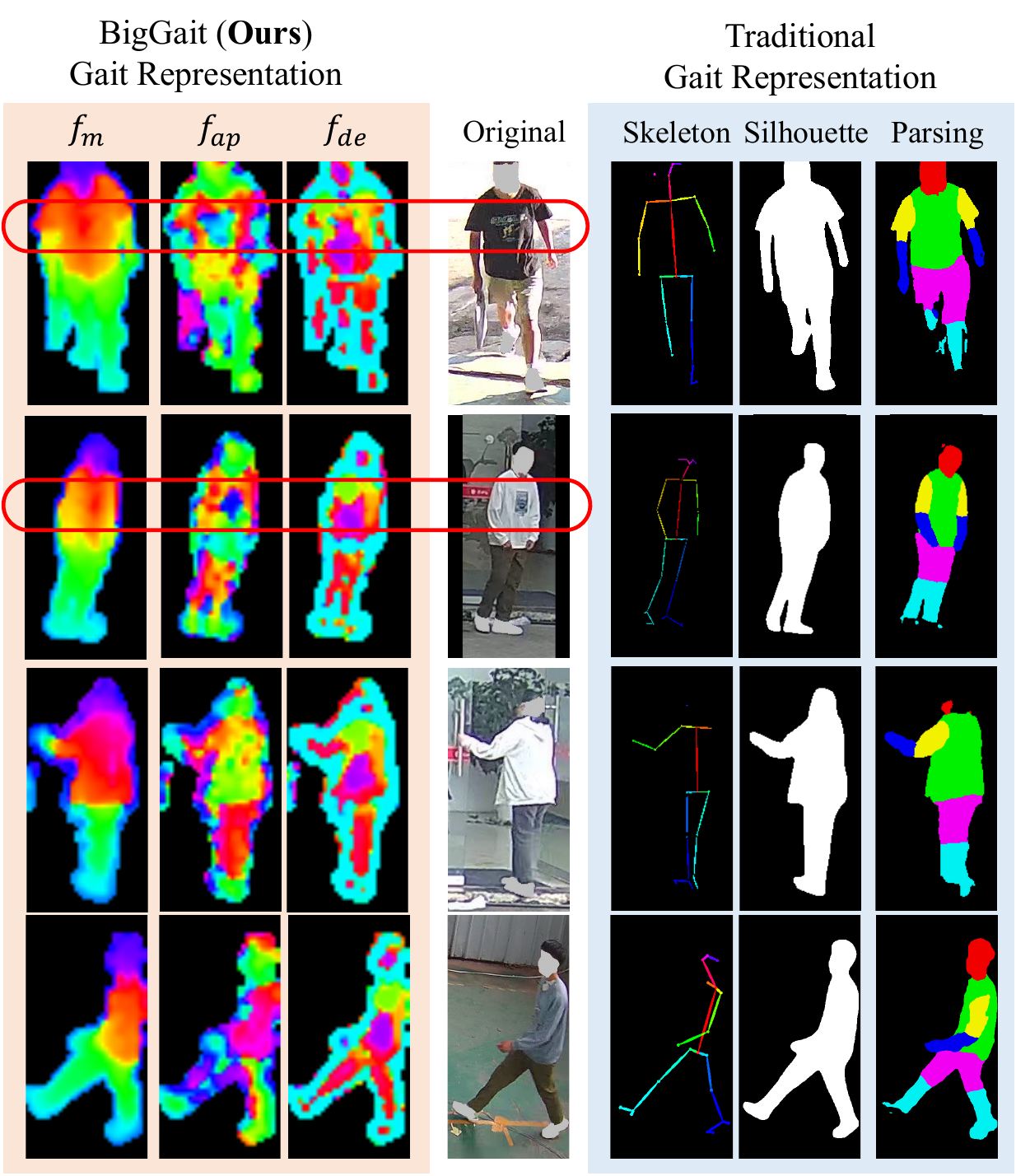}
\vspace{-3mm}
\caption{The visualization of intermediate representations generated by BigGait \textit{v.s.} three traditional gait representations. The red boxes indicate regions with strong texture patterns.}
\label{fig:visualization}
\vspace{-2mm}
\end{figure}

To provide an intuitive understanding, we visualize each intermediate representation in BigGait, and compare it with traditional gait representations, as shown in Fig.~\ref{fig:visualization}.
All-purpose features $f_m$ created by the upstream model (DINOv2) provide diverse and discriminative body parts after masking the background regions, \textit{i.e.}, purple head, red abdomen, green limbs and blue shoe.
Both $f_{ap}$ in the appearance branch and $f_{de}$ in the denoising branch are derived from $f_m$, with the former inheriting features by a linear transformation and showing body shape representation with high-frequency texture noises, and the latter embodying highly consistent skeleton-like representation since deploying smoothness and diversity constraints removes most texture noises.
However, compared with traditional gait representations with direct physical meanings, the gait representation based on soft geometrical constraints may need more interpretability on its physical meaning in future.

\subsection{Implementation Details}
\noindent \textbf{Loss.}
The overall loss can be formulated as: 
\vspace{-2.mm}
\begin{equation}
    L = L_{tri} + L_{ce} + \gamma_{rec} L_{rec} + \gamma_{smo} L_{smo} + \gamma_{div} L_{div},
\vspace{-2.mm}
\end{equation}
where the first two terms are recognition losses of the downstream gait model, $L_{rec}$ is the mask reconstruction loss, and the last two are the smoothness and diversity losses of the denoising branch. 
In our implementations, 
$\gamma_{rec}, \gamma_{smo}$ and $\gamma_{div}$ are respectively set to $1.0$, $0.01$, and $5.0$. 

\begin{table}[t]
\centering
\caption{The amount of the identities (\#ID) and sequences (\#Seq) covered by the employed datasets.}
\vspace{-2.mm}
\renewcommand{\arraystretch}{0.85}
\resizebox{0.9\columnwidth}{!}{

\begin{tabular}{cccccc} 
\toprule[2pt]
\multirow{2}{*}{Data Set} & \multicolumn{2}{c}{Train Set} & \multicolumn{2}{c}{Test Set} & \multirow{2}{*}{Setting}  \\
                         & \#ID & \#Seq                  & \#ID & \#Seq                 &                             \\ 
\midrule[1pt]
CCPG                     & $100$  & $8,388$                  & $100$  & $8,178$                 & CL, UP, DN, BG                 \\
CASIA-B*                 & $74$   & $8,140$                  & $50$   & $5,500$                 & NM, BG, CL                    \\
SUSTech1K                & $250$  & $6,011$                  & $750$  & $19,228$                & Various                     \\
\bottomrule[2pt]
\end{tabular}

}
\label{tab: dataset}
\vspace{-5mm}
\end{table}

\noindent \textbf{Training.}
All images are processed with Pad-and-Resize to the size of $448\times224$.
We use the SGD optimizer with a weight decay of $0.0005$, a momentum of $0.9$ and an initial learning rate of $0.1$.
The learning rate is reduced by a factor of $0.1$ at the $15$k, $25$k, $30$k and $35$k iterations, and the total number of iterations is  $40$k.
The batch size $(p,k,l)$ is set to $(8,8,30)$, where $p$ is the number of IDs, $k$  the number of sequences per ID, and $l$ the number of frames per sequence.
The sampling strategy of frames from sequence follows GaitBase~\cite{fan2022opengait}.
The data augmentation operation only contains randomly flipping all images in a sequence.

\begin{table}[t]
\centering
\caption{
Rank-1 accuracy on CCPG: BigGait \textit{v.s.}~recent SoTA models using different inputs. 
$\text{GaitBase}^{p}$: human parsing inputs. $\text{GaitBase}^{p+s}$: parsing and silhouette inputs. 
}
\vspace{-2mm}
\renewcommand{\arraystretch}{1.2}
\large
\resizebox{1\columnwidth}{!}{
\begin{tabular}{c|c|c|cccc|c} 
\toprule[2pt]
Input                     & Model                                               & Venue                                        & CL                                       & UP                                       & DN                                       & BG                                       & Mean           \\ 
\midrule
\multirow{3}{*}{Skeleton} & GaitGraph2~\cite{teepe2022towards} & CVPRW'22 & $5.0$                                      & $5.3$                                      & $5.8$                                      & $6.2$                                      & $5.6$            \\
                          & Gait-TR~\cite{zhang2023spatial}    & ES'23                                        & $15.7$                                     & $18.3$                                     & $18.5$                                     & $17.5$                                     & $17.5$           \\
                          & MSGG~\cite{peng2023learning}       & MTA'23                                       & $29.0$                                     & $34.5$                                     & $37.1$                                     & $33.3$                                     & $33.5$           \\ 
\midrule
\multirow{4}{*}{Sils}     & GaitSet~\cite{Chao2019}            & TPAMI'22                                     & $60.2$                                     & $65.2$                                     & $65.1$                                     & $68.5$                                     & $64.8$           \\
                          & GaitPart~\cite{fan2020gaitpart}    & CVPR'20                                      & $64.3$                                     & $67.8$                                     & $68.6$                                     & $71.7$                                     & $68.1$           \\
                          & AUG-OGBase~\cite{li2023depth}      & CVPR'23                                      & $52.1$                                     & $57.3$                                     & $60.1$                                     & $63.3$                                     & $58.2$           \\
                          & GaitBase~\cite{fan2022opengait}    & CVPR'23                                      & $71.6$                                     & $75.0$                                     & $76.8$                                     & $78.6$                                     & $75.5$           \\ 
                          & DeepGaitV2~\cite{fan2023exploring}        & Arxiv                                        & $78.6$          & $84.8$          & $80.7$          & $89.2$          & $83.3$           \\ 

\midrule
Parsing                   & $\text{GaitBase}^p$                                 & CVPR'23                                      & $59.1$                                     & $62.1$                                     & $66.8$                                     & $68.1$                                     & $64.0$           \\ 
\midrule
Parsing+Sils              & $\text{GaitBase}^{p+s}$                               & CVPR'23                                      & $73.6$                                     & $76.2$                                     & $79.1$                                     & $79.2$                                     & $77.0$           \\ 
\midrule
Skeleton+Sils             & SkeletonGait++~\cite{fan2023skeletongait} & AAAI'24                                      & $79.1$          & $83.9$          & $81.7$          & $89.9$          & $83.7$           \\ 
\midrule
RGB+Sils                  & GaitEdge~\cite{liang2022gaitedge}  & ECCV'22                                      & $66.9$                                     & $74.0$                                     & $70.6$                                     & $77.1$                                     & $72.2$           \\ 
\midrule
\multirow{4}{*}{RGB}      & AP3D~\cite{gu2020appearance}       & ECCV'20                                      & $53.4$                                     & $57.3$                                     & $69.7$                                     & $91.4$                                     & $67.8$           \\
                          & PSTA~\cite{wang2021pyramid}        & ICCV'21                                      & $42.2$                                     & $52.2$                                     & $60.3$                                     & $84.5$                                     & $59.8$           \\
                          & PiT~\cite{zang2022multidirection}  & TII'22                                       & $41.0$                                     & $47.6$                                     & $64.3$                                     & $91.0$                                     & $61.0$           \\ 
\cmidrule{2-8}
                          & BigGait & \multirow{1}{*}{CVPR'24} & $\mathbf{82.6}$ & $\mathbf{85.9}$ & $\mathbf{87.1}$ & $\mathbf{93.1}$ & $\mathbf{87.2}$  \\
\bottomrule[2pt]
\end{tabular}
}
\label{tab: recognition performance}
\vspace{-5mm}
\end{table}
\begin{table*}[t]
\centering
\caption{
Rank-1 accuracy on three popular benchmark datasets: BigGait \textit{v.s.}~recent SoTA methods.
This is a cross-domain evaluation and comparison, in which all methods are trained on one dataset and tested on the remaining two datasets.
}
\vspace{-0.5em}
\renewcommand{\arraystretch}{1.2}

\resizebox{2.0\columnwidth}{!}{

\fontsize{30}{35}\selectfont

\begin{tabular}{c|ccc|ccc}
\multicolumn{7}{c}{\fontsize{35}{35}\selectfont (a) Trained on \textbf{CCPG}}                                                                                                                            \\ 
\toprule[5pt]
\multirow{3}{*}{Model}                           & \multicolumn{6}{c}{Test Set}                                                                                              \\ 
\cline{2-7}
                                                 & \multicolumn{3}{c|}{CASIA-B*}                 & \multicolumn{3}{c}{SUSTech1K}                                             \\ 
\cline{2-7}
                                                 & NM            & BG            & CL            & Clothing               & Night                  & Overall                 \\ 
\midrule
GaitSet~\cite{Chao2019}                                          & \scalebox{1.2}{$47.4$}          & \scalebox{1.2}{$40.9$}          & \scalebox{1.2}{$25.8$}          & \scalebox{1.2}{$8.2$}                    & \scalebox{1.2}{$11.0$}                   & \scalebox{1.2}{$12.8$}                    \\
GaitBase~\cite{fan2022opengait} & \scalebox{1.2}{$59.1$}          & \scalebox{1.2}{$52.7$}          & \scalebox{1.2}{$30.4$}          & \scalebox{1.2}{$9.5$}                    & \scalebox{1.2}{$13.1$}                   & \scalebox{1.2}{$16.8$}                    \\
AP3D~\cite{gu2020appearance}    & \scalebox{1.2}{$53.7$}          & \scalebox{1.2}{$46.2$}          & \scalebox{1.2}{$11.9$}          & \scalebox{1.2}{$36.2$}                   & \scalebox{1.2}{$\mathbf{51.6}$}                   & \scalebox{1.2}{$55.3$}                    \\
PSTA~\cite{wang2021pyramid}                                             & \scalebox{1.2}{$49.7$}          & \scalebox{1.2}{$42.3$}          & \scalebox{1.2}{$8.8$}           & \scalebox{1.2}{$25.7$}                   & \scalebox{1.2}{$29.4$}                   & \scalebox{1.2}{$40.6$}                    \\
BigGait                                          & \scalebox{1.2}{$\mathbf{77.4}$} & \scalebox{1.2}{$\mathbf{71.5}$} & \scalebox{1.2}{$\mathbf{33.6}$} & \scalebox{1.2}{$\mathbf{43.7}$} & \scalebox{1.2}{$44.8$} & \scalebox{1.2}{$\mathbf{56.4}$}  \\
\bottomrule[5pt]
\end{tabular}

\quad

\begin{tabular}{c|cccc|ccc}
\multicolumn{8}{c}{\fontsize{35}{35}\selectfont (b) Trained on \textbf{CASIA-B*}}                                                                                                              \\ 
\toprule[5pt]
\multirow{3}{*}{Model} & \multicolumn{7}{c}{Test Set}                                                                                                              \\ 
\cline{2-8}
                       & \multicolumn{4}{c|}{CCPG}                                     & \multicolumn{3}{c}{SUSTech1K}                                             \\ 
\cline{2-8}
                       & CL            & UP            & DN            & BG            & Clothing               & Night                  & Overall                 \\ 
\midrule
GaitSet                & \scalebox{1.2}{$\mathbf{10.6}$} & \scalebox{1.2}{$16.4$}          & \scalebox{1.2}{$17.2$}          & \scalebox{1.2}{$24.9$}          & \scalebox{1.2}{$7.2$}                    & \scalebox{1.2}{$12.2$}                   & \scalebox{1.2}{$12.8$}                    \\
GaitBase               & \scalebox{1.2}{$\mathbf{10.6}$} & \scalebox{1.2}{$18.1$}          & \scalebox{1.2}{$\mathbf{21.4}$} & \scalebox{1.2}{$28.7$}          & \scalebox{1.2}{$8.1$}                    & \scalebox{1.2}{$11.8$}                   & \scalebox{1.2}{$15.6$}                    \\
AP3D                   & \scalebox{1.2}{$2.1$}           & \scalebox{1.2}{$2.9$}           & \scalebox{1.2}{$3.9$}           & \scalebox{1.2}{$6.1$}           & \scalebox{1.2}{$29.3$}                   & \scalebox{1.2}{$47.4$}                   & \scalebox{1.2}{$48.3$}                    \\
PSTA                   & \scalebox{1.2}{$1.7$}           & \scalebox{1.2}{$1.9$}           & \scalebox{1.2}{$3.4$}           & \scalebox{1.2}{$5.0$}           & \scalebox{1.2}{$19.9$}                   & \scalebox{1.2}{$37.5$}                   & \scalebox{1.2}{$34.6$}                    \\
BigGait                & \scalebox{1.2}{$7.5$}           & \scalebox{1.2}{$\mathbf{19.5}$} & \scalebox{1.2}{$14.2$}          & \scalebox{1.2}{$\mathbf{43.0}$} & \scalebox{1.2}{$\mathbf{36.9}$} & \scalebox{1.2}{$\mathbf{60.2}$} & \scalebox{1.2}{$\mathbf{64.8}$}  \\
\bottomrule[5pt]
\end{tabular}

\quad

\begin{tabular}{c|cccc|ccc}
\multicolumn{8}{c}{\fontsize{35}{35}\selectfont (c) Trained on \textbf{SUSTech1K}}                   \\ 
\toprule[5pt]
\multirow{3}{*}{Model} & \multicolumn{7}{c}{Test Set}                                                                                   \\ 
\cline{2-8}
                       & \multicolumn{4}{c|}{CCPG}                                     & \multicolumn{3}{c}{CASIA-B*}                   \\ 
\cline{2-8}
                       & CL            & UP            & DN            & BG            & NM            & BG            & CL             \\ 
\midrule
GaitSet                & \scalebox{1.2}{$14.0$}             & \scalebox{1.2}{$\mathbf{23.7}$}             & \scalebox{1.2}{$20.3$}             & \scalebox{1.2}{$43.2$}             & \scalebox{1.2}{$63.3$}             & \scalebox{1.2}{$50.8$}             & \scalebox{1.2}{$26.4$}              \\
GaitBase               & \scalebox{1.2}{$\mathbf{16.8}$} & \scalebox{1.2}{$21.7$} & \scalebox{1.2}{$\mathbf{26.0}$} & \scalebox{1.2}{$42.7$}          & \scalebox{1.2}{$73.1$}          & \scalebox{1.2}{$61.2$}          & \scalebox{1.2}{$\mathbf{28.2}$}  \\
AP3D                   & \scalebox{1.2}{$5.5$}           & \scalebox{1.2}{$7.9$}           & \scalebox{1.2}{$13.9$}          & \scalebox{1.2}{$35.1$}          & \scalebox{1.2}{$56.7$}          & \scalebox{1.2}{$48.1$}          & \scalebox{1.2}{$15.3$}           \\
PSTA                   & \scalebox{1.2}{$3.7$}           & \scalebox{1.2}{$5.7$}           & \scalebox{1.2}{$9.5$}           & \scalebox{1.2}{$26.5$}          & \scalebox{1.2}{$31.2$}          & \scalebox{1.2}{$27.7$}          & \scalebox{1.2}{$10.6$}           \\
BigGait                & \scalebox{1.2}{$4.5$}           & \scalebox{1.2}{$11.5$}          & \scalebox{1.2}{$11.9$}          & \scalebox{1.2}{$\mathbf{45.5}$} & \scalebox{1.2}{$\mathbf{91.1}$} & \scalebox{1.2}{$\mathbf{85.8}$} & \scalebox{1.2}{$18.7$}           \\
\bottomrule[5pt]
\end{tabular}
}
\label{tab: Generalization Performance}
\vspace{-1em}

\end{table*}

\section{Experiments}
\label{sec:experiments}

\subsection{Datasets}
In our experiments, three popular cross-clothing and multi-view gait datasets are employed, \textit{i.e.}, CCPG~\cite{li2023depth}, CASIA-B*~\cite{yu2006framework, liang2022gaitedge} and SUSTech1K~\cite{Shen_2023_CVPR}.
Among them, CCPG acts as the primary benchmark since it prioritizes challenging cloth-changing scenarios, featuring a diverse collection of coats, pants, and bags in various colors and styles. 
The key statistics of these gait datasets are listed in Tab.~\ref{tab: dataset}.
Our implementation follows official protocols, including the training and gallery/probe set partition strategies.
During the testing, all datasets adopt comprehensive gait evaluation protocols for multi-view scenes, and rank-1 accuracy is considered the primary metric.


\vspace{-1.mm}
\subsection{Main Results}
Note BigGait stands for BigGait-S, as mentioned in Sec.~\ref{sec:method_overview}, where BigGait-L refers to its large version. 

\noindent \textbf{Within-domain Evaluation.} 
We compare BigGait with various SoTA methods, such as the skeleton-based~\cite{teepe2022towards,zhang2023spatial,peng2023learning}, silhouette-based~\cite{Chao2019,fan2020gaitpart,li2023depth,fan2022opengait}, and end-to-end gait recognition methods~\cite{liang2022gaitedge}, and video-based ReID methods~\cite{gu2020appearance,wang2021pyramid,zang2022multidirection}, on the challenging CCPG.
The Supplementary Material provides more within-domain results on ReID evaluation protocols and on other datasets~\cite{yu2006framework,Shen_2023_CVPR,Zou2024CCGR}.

As shown in Tab.~\ref{tab: recognition performance}, BigGait achieves significantly better performance than other SoTA methods on CCPG.
In the following, we measure BigGait's effectiveness in filtering out gait-irrelevant noises from two aspects.

\begin{figure}[t]
\centering
    \includegraphics[width=1\columnwidth]{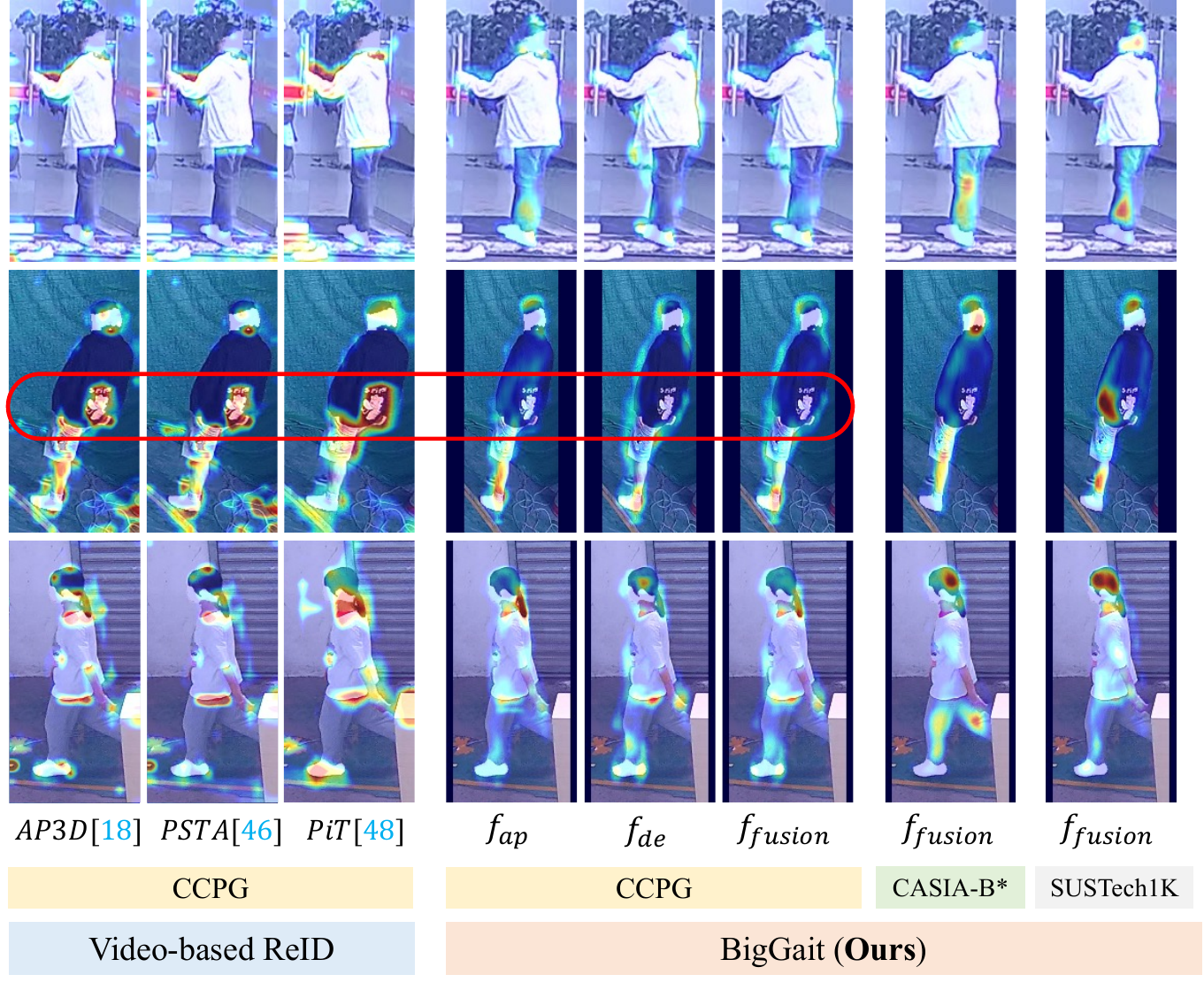}
    \vspace{-1.5em}
    
\caption{
Visualization of activation maps. 
By background removal and texture denoising, BigGait pay more attention on robust gait pattern than video-based ReID methods. 
}
\label{fig: heatmap}
\vspace{-1em}
\end{figure}

First, compared to video-based ReID methods~\cite{gu2020appearance,wang2021pyramid,zang2022multidirection}, BigGait outperforms them considerably, \textit{e.g.}, $+29.2\%$ for full-changing (CL), $+28.6\%$ for ups-changing (UP), $+17.4\%$ for pants-changing (DN) and $+1.7\%$ for bag-changing (BG) scenarios.
Accordingly, we consider that BigGait can efficiently extract robust identity representations from raw RGB videos.
Fig.~\ref{fig: heatmap} provides intuitive evidence by visualizing the activation difference between BigGait and video-based ReID methods.

Second, compared to silhouette-based gait methods~\cite{Chao2019,fan2020gaitpart,li2023depth,fan2022opengait}, which perform well due to immunity to color/texture noises, BigGait still outperforms them by $+4.0\%$, $+1.1\%$, $+6.4\%$, and $+3.9\%$ in CL, UP, DN, and BG settings. 
This indicates that BigGait's immunity to gait-irrelevant noises is comparable to silhouette-based ones.

Based on these analyses, we consider that gait features are the primary characteristics extracted by BigGait, even if gait-irrelevant noises cannot be absolutely eliminated from RGB inputs. 
We further validate this conclusion through cross-domain and visualization experiments.

\noindent \textbf{Cross-domain Evaluation.} 
Tab.~\ref{tab: Generalization Performance} shows cross-domain performance comparisons.
BigGait's cross-domain performance varies depending on the training set used.
When trained on CCPG, BigGait exhibits strong adaptability to unseen datasets, \textit{i.e.}, outperforming both video-based ReID methods~\cite{gu2020appearance,wang2021pyramid} and silhouette-based methods~\cite{fan2022opengait,Chao2019}.
When the training set is CASIA-B*, BigGait exihibits less impressive performances in some settings, \textit{i.e.}, performing well in the ups-changing (UP), but poorly in the full-changing (CL) and pants-changing (DN).
When crossing domains from SUSTech1K, BigGait struggles in more cross-dressing settings (CL, UP and DN).

We consider that BigGait's power in filtering gait-irrelevant noises is impacted by the training data biases.
Specifically, the probability of sampling cross-dressing data pairs from SUSTech1K is lower than $5.0\%$, likely making BigGait focus on unchanged clothing cues rather than subtle gait patterns.
Fortunately, with the enrichment of training data, BigGait can self-correct this mistake.
CASIA-B* includes over $35.0\%$ cross-dressing sample pairs, all of which are ups-changing, which improve BigGait's resistance to ups-changing (UP), as shown in Tab.~\ref{tab: Generalization Performance} (b).
CCPG has more diverse outfits than others, containing over $95.0\%$ cross-dressing pairs, with $78.5\%$ for ups-changing, $80.0\%$ for pants-changing, and $62.0\%$ for full-changing.
As shown in  Tab.~\ref{tab: Generalization Performance} (a), BigGait can learn robust gait representations by training on CCPG.
These analyses indicate that the distribution of training data can significantly influence BigGait's outcomes, 
\textit{i.e}, the more cross-dressing changes given, the more cross-dressing capacity obtained.

\noindent \textbf{Visualization of Activation Maps.} 
To delve deeper into the gait representation learned by BigGait, Fig.~\ref{fig: heatmap} shows visualizations of activation maps from both BigGait and video-based ReID methods~\cite{gu2020appearance,wang2021pyramid,zang2022multidirection}, driven by the popular Grad-CAM~\cite{selvaraju2017grad} algorithm.
These maps are from the first layer in BigGait's downstream model, second layer in AP3D~\cite{gu2020appearance} and PSTA~\cite{wang2021pyramid}, and fourth block in PiT~\cite{zang2022multidirection}.
Four key insights emerge:
a) Unlike video-based ReID methods distracted by cluttered backgrounds, BigGait solely attends to the body.
b) The video-based ReID methods are sensitive to high-frequency clothing texture, as shown in the red box of Fig.~\ref{fig: heatmap}.
c) $f_{ap}$ and $f_{de}$ target different body regions, with $f_{ap}$ emphasizing limbs and head via parsing-like priors, and $f_{de}$ highlighting human contours by suppressing texture noises on body.
Fusing $f_{ap}$ and $f_{de}$ using attention weights to obtain $f_{fusion}$.
d) As mentioned before, CASIA-B* only contains ups-changes, and SUSTech1K has few cloth-changes.
Thus BigGait incorrectly highlights unchanged pants and clothing when trained on CAISA-B* and SUSTech1K, respectively.
More visualizations are provided in Supplementary Material.

\begin{table}[t]
\centering
\caption{Ablation of each branch and Pad-and-Resize: w/ and w/o mask branch, denoising branch, appearance branch and Pad-and-Resize strategy. \checkmark* denotes direct feeding to the downstream.}
\vspace{-1.em}
\renewcommand{\arraystretch}{1}
\resizebox{0.95\columnwidth}{!}{

\begin{tabular}{c|cccc|cccc} 
\toprule[2pt]
Index                         & Mask                       & \begin{tabular}[c]{@{}c@{}}Denois-\\ing\end{tabular} & \begin{tabular}[c]{@{}c@{}}Appear-\\ance\end{tabular} & \begin{tabular}[c]{@{}c@{}}Pad-and-\\Resize\end{tabular} & CL                                       & UP                                       & DN                                       & BG             \\ 
\midrule[1pt]
\rowcolor[rgb]{1,1,0.784} (a) & \checkmark  & \checkmark                            & \checkmark                             & \checkmark                                & $\mathbf{76.0}$ & $\mathbf{79.1}$ & $\mathbf{84.2}$ & $93.0$           \\
(b)                           & \checkmark  & \checkmark                            & \checkmark                             & \texttimes                                & $70.5$                                     & $72.9$                                     & $82.8$                                     & $93.0$           \\
(c)                           & \texttimes  & \texttimes                            & \texttimes                             & \checkmark                                & $55.6$                                     & $62.7$                                     & $73.7$                                     & $87.8$           \\
(d)                           & \texttimes  & \checkmark                            & \checkmark                             & \checkmark                                & $70.5$                                     & $76.1$                                     & $80.9$                                     & $92.2$           \\
(e)                           & \checkmark  & \texttimes                            & \checkmark                             & \checkmark                                & $66.1$                                     & $71.6$                                     & $80.7$                                     & $\mathbf{93.2}$  \\
(f)                           & \checkmark  & \checkmark                            & \texttimes                             & \checkmark                                & $65.1$                                     & $68.4$                                     & $79.2$                                     & $90.1$           \\
(g)                           & \checkmark* & \texttimes                            & \texttimes                             & \checkmark                                & $36.7$                                     & $46.8$                                     & $52.4$                                     & $46.4$           \\
\bottomrule[2pt]
\end{tabular}

}
\vspace{-1em}
\label{tab: Each Branch}
\end{table}

\subsection{Ablation Study} 
All experiments are conducted on CCPG.
Due to limited GPU resources, we adopt $8$ frames per sequence in the ablation study to train BigGait for more efficient iterations.
Despite slight performance drop, BigGait's performance in Tab.~\ref{tab: Each Branch} (a) remains comparable with SoTA methods in Tab.~\ref{tab: recognition performance}.

\noindent \textbf{Pad-and-Resize.} 
Compared (a) and (b) in Tab.~\ref{tab: Each Branch}, the Pad-and-Resize strategy is helpful for BigGait by faithfully preserving body aspect ratio.

\noindent \textbf{Various Branch.} 
In Tab.~\ref{tab: Each Branch} (c), without the GRE module, the performance of BigGait drops to be similar to video-based ReID methods~\cite{gu2020appearance} in Tab.~\ref{tab: recognition performance}.
This highlights the effectiveness of GRE module.
Compared (a) to (d-f) in Tab.~\ref{tab: Each Branch}, each branch contributes substantially.
This implies a mutually beneficial relationship among them.
In Tab.~\ref{tab: Each Branch} (g), BigGait's foreground mask is less discriminable for recognition compared to the silhouette in Tab.~\ref{tab: recognition performance}.
We consider that BigGait's foreground masks is approximate regions rather than precise segmentations.
SoTA results in Tab.~\ref{tab: Each Branch} (a) are achieved even without finely segmented foregrounds.

\noindent \textbf{Denoising Branch.} 
Comparing (a) to (b-d) in Tab.~\ref{tab: Denoising Branch}, smoothness loss alone leads to a trivial solution, while combining it with diversity constraint improves results.
Compared (a) with (e) and (f) in Tab.~\ref{tab: Denoising Branch}, we explore an appropriate channel number for the denoising branch.

\noindent \textbf{Upstream and Downstream Models.} 
Compared (c) with (d) in Tab.~\ref{tab: Upstream Model}, BigGait-L exhibits slightly higher performance than BigGait-S.
Compared (c) with (e) in Tab.~\ref{tab: Upstream Model}, we verify that the gains primarily come from the knowledge embedded in DINOv2 rather than its model architecture.
Compared (c) and (f) in Tab.~\ref{tab: Upstream Model}, the performance suffers when fine-tuning the upstream DINOv2.
Further, we explore the influence of different upstream/downstream models for BigGait.
Compared (c) and (g) in Tab.~\ref{tab: Upstream Model}, although the performance drops when changing BigGait's downstream model, it still surpasses its silhouette-based counterpart in (a).
As shown in Tab.~\ref{tab: Upstream Model} (h), BigGait remains effective even when changing BigGait's upstream model, \textit{i.e.}, replacing DINOv2~\cite{oquab2023dinov2} with SAM~\cite{kirillov2023segment}.
In most cases, it still outperforms its silhouette-based counterpart in (b).
This highlights BigGait's flexibility for upstream/downstream models.

\vspace{-1mm}
\section{Challenges and Limitations}
\label{sec:limitations}
\vspace{-1mm}
\noindent \textbf{Challenges for LVMs-based Gait Recognition.}
Beyond attractive profits brought by BigGait, this paper also reveals two primary challenges for LVMs-based gait recognition: 
a) \textit{Interpretability.} 
While we have introduced several explicit human gait priors as soft constraints, the learned gait representations, in comparison to conventional ones defined by clear and intuitive physical attributes, remain partially understandable. 
b) \textit{Purity.}
In gait methods that directly utilize RGB videos as input, a recurring challenge involves effectively reducing gait-irrelevant noise within walking sequences. 
This task becomes even more demanding when attempting to preserve the purity of gait characteristics in LVMs-based gait recognition without explicit supervision.

\begin{table}[t]
\centering
\caption{Ablation of denoising branch: exploring channel number, and w/ and w/o smoothness and diversity loss.}
\vspace{-1.em}
\renewcommand{\arraystretch}{1}
\resizebox{0.95\columnwidth}{!}{

\begin{tabular}{c|ccc|cccc} 
\toprule[2pt]
Index                         & Dimension & Smoothness                & Diversity                 & CL                                       & UP            & DN                                       & BG             \\ 
\midrule[1pt]
\rowcolor[rgb]{1,1,0.784} (a) & 16        & \checkmark & \checkmark & $\mathbf{76.0}$ & $79.1$          & $\mathbf{84.2}$ & $93.0$           \\
(b)                           & 16        & \texttimes & \texttimes & $69.1$                                     & $73.7$          & $81.3$                                     & $91.4$           \\
(c)                           & 16        & \checkmark & \texttimes & $68.8$                                     & $73.4$          & $80.7$                                     & $92.6$           \\
(d)                           & 16        & \texttimes & \checkmark & $74.5$                                     & $\mathbf{80.1}$ & $83.1$                                     & $93.2$           \\
(e)                           & 8         & \checkmark & \checkmark & $72.8$                                     & $77.3$          & $83.8$                                     & $\mathbf{93.5}$  \\
(f)                           & 32        & \checkmark & \checkmark & $68.5$                                     & $73.4$          & $81.2$                                     & $92.0$           \\
\bottomrule[2pt]
\end{tabular}
}
\vspace{-0.5em}
\label{tab: Denoising Branch}
\end{table}

\begin{table}[t]
\centering
\caption{Ablation of upstream and downstream models.}
\vspace{-1.em}
\renewcommand{\arraystretch}{1.2}
\resizebox{0.95\columnwidth}{!}{

\begin{tabular}{c|c|cc|cccc} 
\toprule[2pt]
\multicolumn{2}{c|}{Index}                                                 & Upstream                                                                    & Downstream                           & CL            & UP                                 & DN                                 & BG                                  \\ 
\cmidrule{1-8}
\multirow{2}{*}{\rotatebox{90}{-}}                      & (a)                             & \multirow{2}{*}{Segmentation}                                               & GaitSet                              & $60.2$                             & $65.2$                             & $65.1$                             & $68.5$                              \\
                                         & (b)                             &                                                                             & GaitBase                             & $71.6$                             & $75.0$                             & $76.8$                             & $78.6$                              \\ 
\hline
\hline
\multirow{6}{*}{\rotatebox{90}{BigGait (\textbf{Ours})}} & {\cellcolor[rgb]{1,1,0.784}}(c) & {\cellcolor[rgb]{1,1,0.784}}DINOv2$^\text{Frozen}$~(ViT-S/14)               & {\cellcolor[rgb]{1,1,0.784}}GaitBase & {\cellcolor[rgb]{1,1,0.784}}$76.0$ & {\cellcolor[rgb]{1,1,0.784}}$79.1$ & {\cellcolor[rgb]{1,1,0.784}}$84.2$ & {\cellcolor[rgb]{1,1,0.784}}$93.0$  \\
                                         & (d)                             & \begin{tabular}[c]{@{}c@{}}DINOv2$^\text{Frozen}$~(ViT-L/14)\\\end{tabular} & GaitBase                             & $\mathbf{79.0}$                    & $\mathbf{82.3}$                    & $\mathbf{86.7}$                    & $\mathbf{94.5}$                     \\
                                         & (e)                             & DINOv2$^\text{Scratch}$~(ViT-S/14)                                         & GaitBase                             & $36.6$                             & $44.0$                             & $62.4$                             & $79.3$                              \\
                                         & (f)                             & DINOv2$^\text{Fine-tune}$~(ViT-S/14)                                                   & GaitBase                             & $74.8$                             & $78.1$                             & $82.6$                             & $92.7$                              \\
                                         & (g)                             & DINOv2$^\text{Frozen}$~(ViT-S/14)                                                      & GaitSet                              & $63.5$                             & $66.5$                             & $71.7$                             & $79.0$                              \\
                                         & (h)                             & SAM$^\text{Frozen}$~(ViT-B/16)                                                         & GaitBase                             & $71.2$                             & $76.0$                             & $84.8$                             & $93.5$                              \\
\bottomrule[2pt]
\end{tabular}

}
\label{tab: Upstream Model}
\vspace{-1em}
\end{table}

\noindent \textbf{Limitations of this paper.}
a) The influence of different upstream LVMs for BigGait is only preliminarily explored. 
This issue is worth further study. 
b) The GRE module lacks of spatial-temporal designs. 
c) This paper focus on high-frequency texture noises instead of low-frequency color noises, since it is the high-frequency that still heavily impacts existing RGB-based methods in the red box of Fig.~\ref{fig: heatmap}.
Further improvements to color noises are expected.

\vspace{-1mm}
\section{Conclusion}
\label{sec:conclusion}
\vspace{-1mm}


In this paper, we present an innovative and efficient methodology for the next-generation gait representation construction, termed BigGait, with gait guidance shifting from task-specific priors to LVMs-based all-purpose knowledge.
Results on CCPG, CASIA-B* and SUSTech1K show that BigGait significantly outperforms the previous methods in both within-domain and cross-domain tasks in most cases, indicating that BigGait is a more practical paradigm for learning general gait representation.
Moreover, this work discusses the challenge in LVMs-based gait recognition, which provides some possible directions on future research.
The work may also provide inspiration to employ all-purpose knowledge produced by LVMs for other vision tasks.

\vspace{-1mm}
\section*{Acknowledgements}
\vspace{-1mm}
This work was supported in part by Shenzhen International Research Cooperation Project (Grant No. GJHZ20220913142611021) and National Natural Science Foundation of China (Grant No. 61976144).

\newpage

\appendix

\twocolumn[{ %
\renewcommand\twocolumn[1][]{#1} %
\begin{center}
    \title{BigGait: Learning Gait Representation You Want by Large Vision Models\vspace{-4em}}  
    \maketitle
    \captionsetup{type=figure}
    \centering
    \includegraphics[width=2\columnwidth]{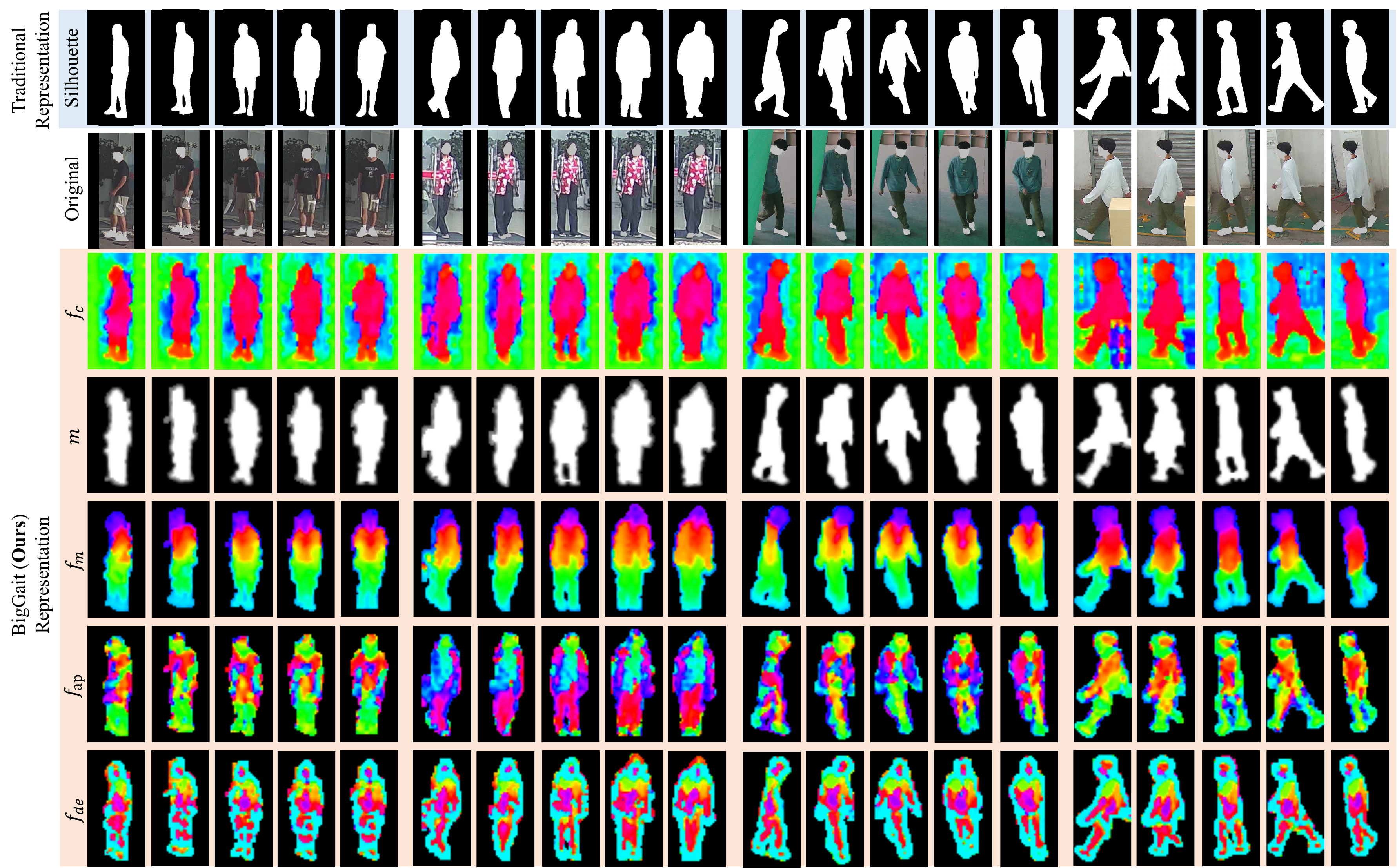}
    \caption{Visualization of intermediate representations. This figure is a supplement for Fig.~5.}
    \label{fig: sup_0}
\end{center}
}]

\begin{table*}[t]
\centering
\caption{\textbf{Within-domain Evaluation.} Rank-1 accuracy on the four popular benchmark datasets under the within-domain task: BigGait \textit{v.s.} recent SoTA methods.
}
\renewcommand{\arraystretch}{1.1}
\resizebox{2.05\columnwidth}{!}{

\begin{tabular}{c|c|cccc|ccc|cccccc|cccc} 
\toprule[2pt]
\multirow{3}{*}{Type}                                                       & \multirow{3}{*}{Model}                           & \multicolumn{17}{c}{Within-domian Evaluation}                                                                                                                                                                                                                                                                                                                                                         \\ 
\cline{3-19}
                                                                            &                                                  & \multicolumn{4}{c|}{CCPG~\cite{li2023depth}}         & \multicolumn{3}{c|}{CASIA-B*~\cite{yu2006framework}} & \multicolumn{6}{c|}{SUSTech1K~\cite{Shen_2023_CVPR}}                                                              & \multicolumn{4}{c}{CCGR~\cite{Zou2024CCGR}}                                                   \\ 
\cline{3-19}
                                                                            &                                                  & CL              & UP              & DN              & BG              & NM               & BG              & CL                               & Normal          & Umbrella        & Uniform         & Clothing                                   & Night           & Overall         & R-1\textsuperscript{hard} & R-1\textsuperscript{easy} & R-5\textsuperscript{hard} & R-5\textsuperscript{easy}  \\ 
\hline
\multirow{2}{*}{\begin{tabular}[c]{@{}c@{}}Gait\\Recognition\end{tabular}}  & GaitSet~\cite{Chao2019}         & \scalebox{1.2}{$60.2$}          & \scalebox{1.2}{$65.2$}          & \scalebox{1.2}{$65.1$}          & \scalebox{1.2}{$68.5$}          & \scalebox{1.2}{$92.3$}           & \scalebox{1.2}{$86.1$}          & \scalebox{1.2}{$73.4$}                           & \scalebox{1.2}{$71.4$}          & \scalebox{1.2}{$66.2$}          & \scalebox{1.2}{$63.8$}          & \scalebox{1.2}{$39.4$}                                     & \scalebox{1.2}{$24.0$}          & \scalebox{1.2}{$67.1$}          & \scalebox{1.2}{$25.3$}                    & \scalebox{1.2}{$35.3$}                    & \scalebox{1.2}{$46.7$}                    & \scalebox{1.2}{$58.9$}                     \\
                                                                            & GaitBase~\cite{fan2022opengait} & \scalebox{1.2}{$71.6$}          & \scalebox{1.2}{$75.0$}          & \scalebox{1.2}{$76.8$}          & \scalebox{1.2}{$78.6$}          & \scalebox{1.2}{$96.5$}           & \scalebox{1.2}{$91.5$}          & \scalebox{1.2}{$78.0$}                           & \scalebox{1.2}{$80.9$}          & \scalebox{1.2}{$74.8$}          & \scalebox{1.2}{$76.3$}          & \scalebox{1.2}{$47.2$}                                     & \scalebox{1.2}{$26.4$}          & \scalebox{1.2}{$75.8$}          & \scalebox{1.2}{$31.3$}                    & \scalebox{1.2}{$43.8$}                    & \scalebox{1.2}{$51.3$}                    & \scalebox{1.2}{$64.4$}                     \\ 
\cline{1-1}
\multirow{2}{*}{\begin{tabular}[c]{@{}c@{}}Video-\\based ReID\end{tabular}} & AP3D~\cite{gu2020appearance}    & \scalebox{1.2}{$53.4$}          & \scalebox{1.2}{$57.3$}          & \scalebox{1.2}{$69.7$}          & \scalebox{1.2}{$91.4$}          & \scalebox{1.2}{$99.8$}           & \scalebox{1.2}{$99.4$} & \scalebox{1.2}{$87.6$}                  & \scalebox{1.2}{$94.4$}          & \scalebox{1.2}{$95.3$}          & \scalebox{1.2}{$91.6$}          & \scalebox{1.2}{$\mathbf{82.7}$} & \scalebox{1.2}{$\mathbf{89.4}$} & \scalebox{1.2}{$\mathbf{96.8}$} & \scalebox{1.2}{$70.2$}                    & \scalebox{1.2}{$82.5$}                    & \scalebox{1.2}{$83.0$}                    & \scalebox{1.2}{$92.4$}                     \\
                                                                            & PSTA~\cite{wang2021pyramid}     & \scalebox{1.2}{$42.2$}          & \scalebox{1.2}{$52.2$}          & \scalebox{1.2}{$60.3$}          & \scalebox{1.2}{$84.5$}          & \scalebox{1.2}{$98.2$}           & \scalebox{1.2}{$96.5$}          & \scalebox{1.2}{$54.2$}                           & \scalebox{1.2}{$92.9$}          & \scalebox{1.2}{$92.1$}          & \scalebox{1.2}{$83.2$}          & \scalebox{1.2}{$72.3$}                                     & \scalebox{1.2}{$79.9$}          & \scalebox{1.2}{$93.6$}          & \scalebox{1.2}{$74.5$}                    & \scalebox{1.2}{$85.0$}                    & \scalebox{1.2}{$86.2$}                    & \scalebox{1.2}{$93.7$}                     \\ 
\cline{1-1}
Ours                                                                        & BigGait                      & \scalebox{1.2}{$\mathbf{82.6}$} & \scalebox{1.2}{$\mathbf{85.9}$} & \scalebox{1.2}{$\mathbf{87.1}$} & \scalebox{1.2}{$\mathbf{93.1}$} & \scalebox{1.2}{$\mathbf{100.0}$} & \scalebox{1.2}{$\mathbf{99.6}$} & \scalebox{1.2}{$\mathbf{90.5}$}                           & \scalebox{1.2}{$\mathbf{96.1}$} & \scalebox{1.2}{$\mathbf{96.0}$} & \scalebox{1.2}{$\mathbf{93.2}$} & \scalebox{1.2}{$73.3$}                                     & \scalebox{1.2}{$85.3$}          & \scalebox{1.2}{$96.2$}          & \scalebox{1.2}{$\mathbf{77.1}$}           & \scalebox{1.2}{$\mathbf{86.2}$}           & \scalebox{1.2}{$\mathbf{87.9}$}           & \scalebox{1.2}{$\mathbf{94.3}$}                     \\
\bottomrule[2pt]
\end{tabular}

}
\label{tab: Self Domain}
\end{table*}

\begin{table*}[t]
\centering
\caption{\textbf{Within-domain Evaluation on CCPG.} This is a supplement for Tab.3, providing more results under ReID Evaluation Protocol.}
\renewcommand{\arraystretch}{1.1}
\resizebox{1.7\columnwidth}{!}{

\begin{tabular}{c|c|c|cccc|c|cccc|c} 
\toprule[2pt]
\multirow{2}{*}{Input}    & \multirow{2}{*}{Model}                                     & \multirow{2}{*}{Venue} & \multicolumn{5}{c|}{Gait Evaluation Protocol}                                                               & \multicolumn{5}{c}{ReID Evaluation Protocol}                                                                 \\ 
\cmidrule{4-13}
                          &                                                            &                        & CL              & UP              & DN              & BG              & Mean            & CL              & UP              & DN              & BG              & Mean             \\ 
\cmidrule{1-13}
\multirow{3}{*}{Skeleton} & GaitGraph2~\cite{teepe2022towards}        & CVPRW'22               & $5.0$           & $5.3$           & $5.8$           & $6.2$           & $5.6$           & $5.0$           & $5.7$           & $7.3$           & $8.8$           & $6.7$            \\
                          & Gait-TR~\cite{zhang2023spatial}           & ES'23                  & $15.7$          & $18.3$          & $18.5$          & $17.5$          & $17.5$          & $24.3$          & $28.7$          & $31.1$          & $28.1$          & $28.1$           \\
                          & MSGG~\cite{peng2023learning}              & MTA'23                 & $29.0$          & $34.5$          & $37.1$          & $33.3$          & $33.5$          & $43.1$          & $52.9$          & $57.4$          & $49.9$          & $50.8$           \\ 
\cmidrule{1-13}
\multirow{5}{*}{Sils}     & GaitSet~\cite{Chao2019}                   & TPAMI'22               & $60.2$          & $65.2$          & $65.1$          & $68.5$          & $64.8$          & $77.5$          & $85.0$          & $82.9$          & $87.5$          & $83.2$           \\
                          & GaitPart~\cite{fan2020gaitpart}           & CVPR'20                & $64.3$          & $67.8$          & $68.6$          & $71.7$          & $68.1$          & $79.2$          & $85.3$          & $86.5$          & $88.0$          & $84.8$           \\
                          & AUG-OGBase~\cite{li2023depth}             & CVPR'23                & $52.1$          & $57.3$          & $60.1$          & $63.3$          & $58.2$          & $70.2$          & $76.9$          & $80.4$          & $83.4$          & $77.7$           \\
                          & GaitBase~\cite{fan2022opengait}           & CVPR'23                & $71.6$          & $75.0$          & $76.8$          & $78.6$          & $75.5$          & $88.5$          & $92.7$          & $93.4$          & $93.2$          & $92.0$           \\
                          & DeepGaitV2~\cite{fan2023exploring}        & Arxiv                  & $78.6$          & $84.8$          & $80.7$          & $89.2$          & $83.3$          & $\mathbf{90.5}$ & $\mathbf{96.3}$ & $91.4$          & $96.7$          & $93.7$           \\ 
\cmidrule{1-13}
Parsing                   & $\text{GaitBase}^p$                                        & CVPR'23                & $59.1$          & $62.1$          & $66.8$          & $68.1$          & $64.0$          & $75.9$          & $81.3$          & $86.5$          & $87.5$          & $82.8$           \\ 
\cmidrule{1-13}
Parsing+Sils              & $\text{GaitBase}^{p+s}$                                      & CVPR'23                & $73.6$          & $76.2$          & $79.1$          & $79.2$          & $77.0$          & $89.3$          & $91.9$          & $93.0$          & $94.3$          & $92.1$           \\ 
\cmidrule{1-13}
Skeleton+Sils             & SkeletonGait++~\cite{fan2023skeletongait} & AAAI'24                & $79.1$          & $83.9$          & $81.7$          & $89.9$          & $83.7$          & $90.2$          & $95.0$          & $92.9$          & $96.9$          & $\mathbf{93.8}$  \\ 
\cmidrule{1-13}
RGB+Sils                  & GaitEdge~\cite{liang2022gaitedge}         & ECCV'22                & $66.9$          & $74.0$          & $70.6$          & $77.1$          & $72.2$          & $73.0$          & $83.5$          & $82.0$          & $87.8$          & $81.6$           \\ 
\cmidrule{1-13}
\multirow{6}{*}{RGB}      & AP3D~\cite{gu2020appearance}              & ECCV'20                & $53.4$          & $57.3$          & $69.7$          & $91.4$          & $67.8$          & $62.6$          & $67.6$          & $82.0$          & $\mathbf{97.3}$          & $77.4$           \\
                          & PSTA~\cite{wang2021pyramid}               & ICCV'21                & $42.2$          & $52.2$          & $60.3$          & $84.5$          & $59.8$          & $51.9$          & $62.0$          & $72.3$          & $94.1$          & $70.1$           \\
                          & PiT~\cite{zang2022multidirection}         & TII'22                 & $41.0$          & $47.6$          & $64.3$          & $91.0$          & $61.0$          & $49.1$          & $56.2$          & $78.0$          & $96.9$          & $70.1$           \\ 
\cmidrule{2-13}
                          & BigGait & \multirow{1}{*}{Ours} & $\mathbf{82.6}$ & $\mathbf{85.9}$ & $\mathbf{87.1}$ & $\mathbf{93.1}$          & $\mathbf{87.2}$ & $89.6$          & $93.2$          & $\mathbf{95.2}$ & $97.2$          & $\mathbf{93.8}$  \\
\bottomrule[2pt]
\end{tabular}

}
\label{tab: ccpg_more}
\end{table*}

\section{Supplementary Material}
\label{sec: Supplementary Material}
In this section, we first provide the details of the Pad-and-Resize trick used for preserving body aspect ratio.
Then more experimental results under both the within and cross-domain scenarios are presented.
Finally, we conducted more visualization to understand the learned characteristics.
Some related issues in rebuttal are attached as well.

\subsection{Pad-and-Resize}
Large vision models tend to directly resize the RGB frame within different bounding boxes into a fixed size, thereby completely or temporarily losing the authenticity of body proportions crucial for gait description.
Comparing (a) with (b) in Tab.5, this lack of attention to body structure can lead to a significant performance drop.
Therefore, this paper adopts a straightforward strategy, \textit{i.e.}, conducting two-side padding or cutting horizontally to force the frame to be a predetermined aspect ratio of $2:1$, and then resizing it into a fixed resolution of $448\times224$, all while maintaining the original aspect ratio of body parts. 
Intuitively, the presence of double-sided black regions shown in Fig.2 (a) contributes to the preservation of limb ratios. 

\subsection{More Experimental Results}

\noindent \textbf{Within-domain Evaluation.}
Tab.~\ref{tab: Self Domain} presents more within-domain performance comparisons.
We observe that BigGait exhibited impressive performance on various kinds of datasets, including CCPG~\cite{li2023depth} with abundant clothing variations, CASIA-B*~\cite{yu2006framework} focusing ups-clothing changes, and nearly cloth-unchanging SUSTech1K~\cite{Shen_2023_CVPR}.
Compared with video-based ReID methods~\cite{gu2020appearance,wang2021pyramid}, BigGait significantly outperforms them on the CCPG~\cite{li2023depth} comprising rich and challenging cloth-changes, and achieves similar performance on other two datasets~\cite{yu2006framework,Shen_2023_CVPR}.
Compared with silhouette-based methods~\cite{fan2022opengait,Chao2019}, BigGait surpasses them on all of these datasets~\cite{li2023depth,yu2006framework,Shen_2023_CVPR}.
Moreover, Tab.~\ref{tab: ccpg_more} provides more within-domain results under ReID evaluation protocols on CCPG.
The above results indicate that BigGait can extract robust gait patterns on different kinds of gait datasets and on different evaluation protocols.

\noindent \textbf{CCGR Evaluation.}
CCGR~\cite{Zou2024CCGR} is a recently released well-labeled gait dataset consisting of over 1.5 million sequences, which has 970 subjects with 33 views and 53 walking conditions.
We evaluate the performance of BigGait trained on CCGR under various tasks.
As shown in Tab.~\ref{tab: Self Domain} and Tab.~\ref{tab: Generalization Performance}, BigGait trained on CCGR surpasses all SoTA methods under the within and cross-domain tasks.
Comparing Tab.~\ref{tab: CCGR Easy Single} and Tab.~\ref{tab: CCGR Easy Mixed}, BigGait presents comparable performance with video-based ReID methods under the single-covariate task, and more outstanding performance under the mixed-covariate task than SoTA methods.
Based on these results, we consider that BigGait learns robust gait representation to resist various covariates.

\subsection{More Visualizations}
\begin{figure*}[ht]
\centering
    \includegraphics[width=0.9\columnwidth]{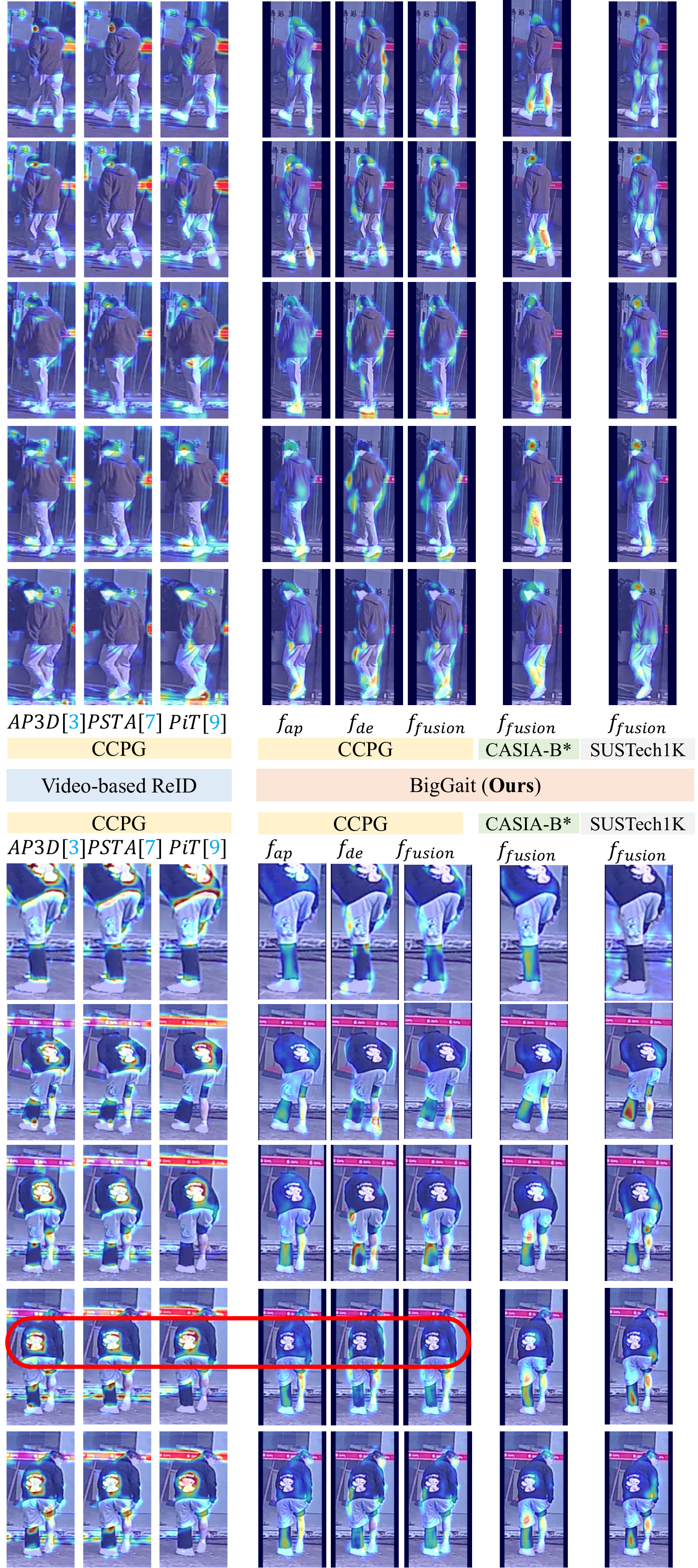}
    \quad
    \includegraphics[width=0.9\columnwidth]{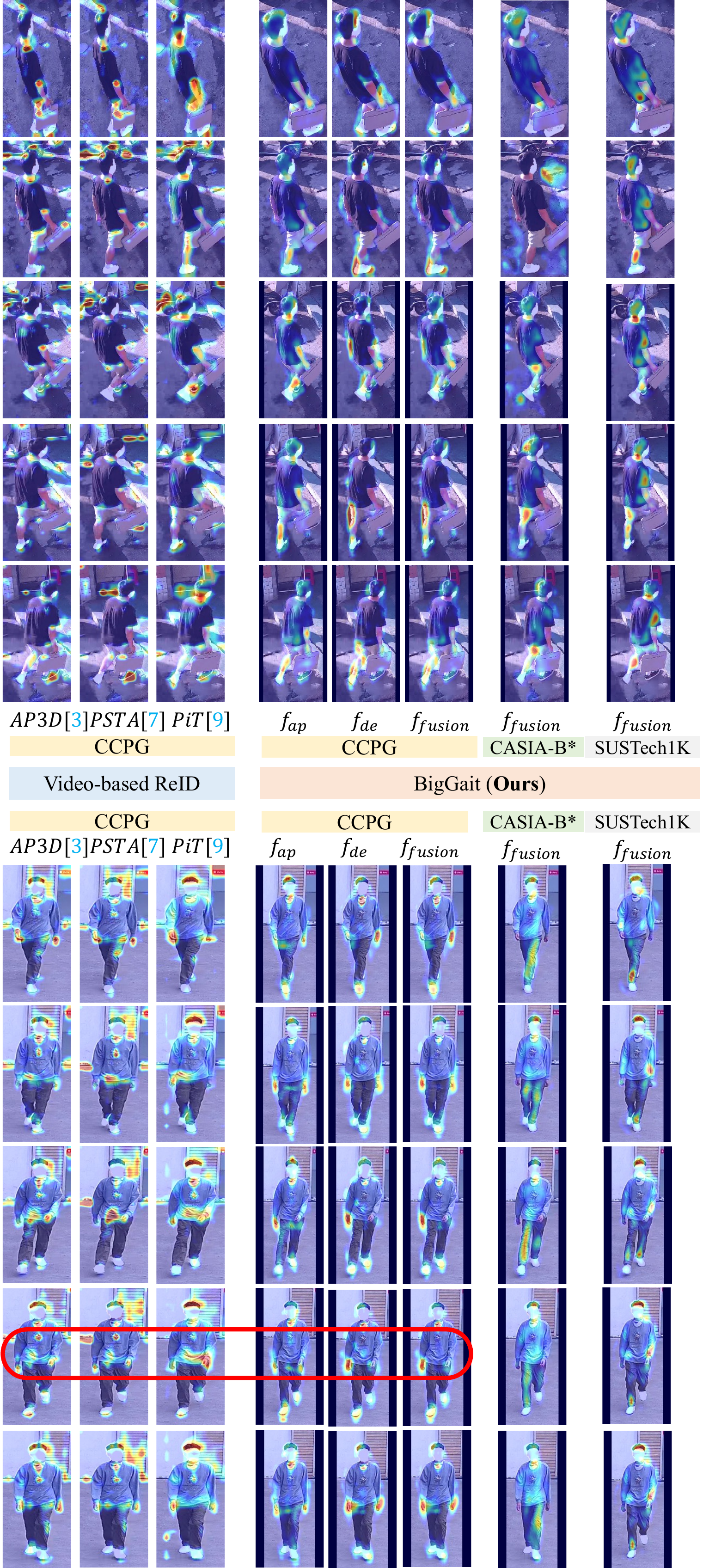}
\caption{
Visualization of activation maps. 
Unlike video-based ReID methods, BigGait focus on robust gait pattern rather than background and clothing texture noises.
This figure is a supplement for Fig.~6.}
\label{fig: sup_1}
\vspace{-1.0em}
\end{figure*}

To better understand the representation learned by BigGait, more visualizations are provided in Fig.~\ref{fig: sup_0} and Fig.~\ref{fig: sup_1}.

\noindent \textbf{Intermediate Feature Maps}. 
Fig.~\ref{fig: sup_0} created by the PCA method exhibits a supplement for Fig.~5 and shows more intermediate feature maps.
As we can see, all-purpose features $f_c$ produced by the upstream DINOv2 are dominated by the separation of foreground and background regions accompanied by noisy spots.
The mask branch in BigGait can automatically infer the foreground mask $m$ from $f_{c}$ in an unsupervised manner.
Compared with silhouettes, $m$ only presents the coarse-grained approximation of body segmentation.
After masking the background regions, $f_{c}$ becomes $f_{m}$ and displays a parsing-like representation, \textit{i.e.}, purple head, red abdomen, yellow arm, green leg, and blue shoe.

However directly using the all-purpose features $f_{m}$ can result in inferior performances, as shown in Tab.~5 (c).
We consider that all-purpose features $f_{m}$ also contain gait-unrelated noise in foreground regions, like the noisy spots in the background regions of $f_{c}$.
To alleviate this problem, the Gait Representation Extractor (GRE) is designed to extract effective gait representations from $f_{m}$ while excluding gait-unrelated noise, as mentioned in Sec.~3.
Specifically, GRE converts $f_{m}$ into $f_{ap}$ and $f_{de}$, with $f_{ap}$ inheriting features by a linear transformation and showing body shape representation with high-frequency texture noise, and $f_{de}$ embodying highly consistent skeleton-like structure representation by deploying soft geometric constraints to denoise most high-frequency texture noise.

\begin{table}[t]
\centering
\caption{\textbf{Cross-domain Evaluation.} This table is a supplement for Tab. 4, in which all methods are trained on CCGR and tested on three unseen datasets.}
\renewcommand{\arraystretch}{1.2}
\resizebox{1.0\columnwidth}{!}{
\begin{tabular}{c|cccc|ccc|ccc}
\toprule[2pt]
\multirow{3}{*}{Model} & \multicolumn{10}{c}{Test Set}                                                                                                                                  \\ 
\cline{2-11}
                       & \multicolumn{4}{c|}{CCPG}                                     & \multicolumn{3}{c|}{CASIA-B*}                 & \multicolumn{3}{c}{SUSTech1K}                  \\ 
\cline{2-11}
                       & CL            & UP            & DN            & BG            & NM            & BG            & CL            & Clothing~     & Night~        & Overall        \\ 
\cline{1-11}
GaitSet                & \scalebox{1.2}{$7.6$}           & \scalebox{1.2}{$13.4$}          & \scalebox{1.2}{$16.2$}          & \scalebox{1.2}{$30.1$}          & \scalebox{1.2}{$32.8$}          & \scalebox{1.2}{$22.2$}          & \scalebox{1.2}{$12.9$}          & \scalebox{1.2}{$13.9$}          & \scalebox{1.2}{$16.9$}          & \scalebox{1.2}{$21.6$}           \\
GaitBase               & \scalebox{1.2}{$5.8$}           & \scalebox{1.2}{$10.2$}          & \scalebox{1.2}{$15.7$}          & \scalebox{1.2}{$26.3$}          & \scalebox{1.2}{$22.1$}          & \scalebox{1.2}{$14.1$}          & \scalebox{1.2}{$7.5$}           & \scalebox{1.2}{$13.4$}          & \scalebox{1.2}{$16.5$}          & \scalebox{1.2}{$27.6$}           \\
AP3D                   & \scalebox{1.2}{$9.8$}           & \scalebox{1.2}{$18.2$}          & \scalebox{1.2}{$25.4$}          & \scalebox{1.2}{$54.7$}          & \scalebox{1.2}{$60.6$}          & \scalebox{1.2}{$55.4$}          & \scalebox{1.2}{$19.7$}          & \scalebox{1.2}{$59.8$}          & \scalebox{1.2}{$48.7$}          & \scalebox{1.2}{$71.3$}           \\
PSTA                   & \scalebox{1.2}{$10.1$}          & \scalebox{1.2}{$17.9$}          & \scalebox{1.2}{$22.0$}          & \scalebox{1.2}{$52.7$}          & \scalebox{1.2}{$31.4$}          & \scalebox{1.2}{$27.8$}          & \scalebox{1.2}{$14.2$}          & \scalebox{1.2}{$55.0$}          & \scalebox{1.2}{$40.3$}          & \scalebox{1.2}{$65.4$}           \\
BigGait                & \scalebox{1.2}{$\mathbf{20.8}$} & \scalebox{1.2}{$\mathbf{38.2}$} & \scalebox{1.2}{$\mathbf{31.9}$} & \scalebox{1.2}{$\mathbf{83.6}$} & \scalebox{1.2}{$\mathbf{93.1}$} & \scalebox{1.2}{$\mathbf{91.8}$} & \scalebox{1.2}{$\mathbf{61.7}$} & \scalebox{1.2}{$\mathbf{73.8}$} & \scalebox{1.2}{$\mathbf{76.8}$} & \scalebox{1.2}{$\mathbf{88.1}$}  \\
\bottomrule[2pt]
\end{tabular}
}
\label{tab: Generalization Performance}
\end{table}

\begin{table}[t]
\centering
\caption{\textbf{Single-Covariate Evaluation}: R-1$^{easy}$ accuracy (\%) with excluding identical-view cases on CCGR dataset.}
\renewcommand{\arraystretch}{1.0}
\resizebox{1.0\columnwidth}{!}{

\begin{tabular}{c|c|c|c|c|c} 
\toprule[2pt]
\multicolumn{6}{c}{Gallery: Normal 1}                                                                                                                       \\ 
\midrule
\multicolumn{3}{c|}{Publication}                                                                    & CVPR'23  & ICCV'21       & Ours  \\ 
\midrule
Type                                                                     & Covariate        & Abbr. & GaitBase & PSTA          & BigGait                    \\ 
\midrule
\multirow{8}{*}{Carrying}                                                & Book             & BK    & $65.7$     & $96.7$          & $94.8$                       \\
                                                                         & Bag              & BG    & $64.9$     & $96.1$          & $94.1$                       \\
                                                                         & Heavy Bag        & HVBG  & $60.0$     & $95.4$          & $93.6$                       \\
                                                                         & Box              & BX    & $61.5$     & $95.6$          & $93.5$                       \\
                                                                         & Heavy Box        & HVBX  & $58.7$     & $95.3$          & $93.6$                       \\
                                                                         & Trolley Case     & TC    & $64.1$     & $94.2$          & $93.0$                       \\
                                                                         & Umbrella         & UB    & $47.2$     & $89.5$          & $85.4$                       \\ 
\cmidrule{2-6}
                                                                         & \textit{Average} & -     & $60.3$     & $\mathbf{94.7}$ & $92.6$                       \\ 
\midrule
Clothing                                                                 & Thick Coat       & CL    & $40.4$     & $88.6$          & $88.7$                       \\ 
\midrule
\multirow{8}{*}{Road}                                                    & Up Ramp          & UTR   & $60.3$     & $90.3$          & $91.7$                       \\
                                                                         & Down Ramp        & DTR   & $60.5$     & $93.2$          & $93.3$                       \\
                                                                         & Up Stair         & UTS   & $54.9$     & $92.0$          & $92.5$                       \\
                                                                         & Down Stair       & DTS   & $54.0$     & $93.3$          & $93.1$                       \\
                                                                         & Bumpy Road       & BM    & $63.3$     & $93.4$          & $93.2$                       \\
                                                                         & Curved Road      & CV    & $70.0$     & $94.4$          & $93.6$                       \\
                                                                         & Soft Road        & SF    & $66.0$     & $93.7$          & $93.1$                       \\ 
\cmidrule{2-6}
                                                                         & \textit{Average} & -     & $61.3$     & $\mathbf{92.9}$ & $\mathbf{92.9}$              \\ 
\midrule
\multirow{4}{*}{Speed}                                                   & Normal 1         & NM1   & $76.6$     & $97.6$          & $96.1$                       \\
                                                                         & Fast             & FA    & $47.2$     & $94.8$          & $91.3$                       \\
                                                                         & Stationary       & ST    & $32.0$     & $92.3$          & $88.1$                       \\ 
\cmidrule{2-6}
                                                                         & \textit{Average} & -     & $51.9$     & $\mathbf{94.9}$ & $91.8$                       \\ 
\midrule
\multirow{5}{*}{\begin{tabular}[c]{@{}c@{}}Walking\\ Style\end{tabular}} & Normal 2         & NM2   & $75.3$     & $97.5$          & $95.8$                       \\
                                                                         & Confident        & CF    & $64.9$     & $96.1$          & $94.3$                       \\
                                                                         & Freedom          & FD    & $57.1$     & $93.7$          & $93.6$                       \\
                                                                         & Multi-person     & MP    & $24.0$     & $51.1$          & $47.8$                       \\ 
\cmidrule{2-6}
                                                                         & \textit{Average} & -     & $55.3$     & $\mathbf{84.6}$ & $82.9$                       \\
\bottomrule[2pt]
\end{tabular}

}
\label{tab: CCGR Easy Single}
\end{table}

\begin{table}[t]
\centering
\caption{\textbf{Mixed-Covariate Evaluation}: R-1$^{easy}$ accuracy (\%) with excluding identical-view cases on CCGR dataset. We use "-" to connect the mixed covariates.}
\renewcommand{\arraystretch}{1.0}
\resizebox{1\columnwidth}{!}{
\begin{tabular}{c|c|c|c|c} 
\toprule[2pt]
\multicolumn{5}{c}{Gallery: Normal 1}                                                                                                                                                                                                   \\ 
\midrule
\multicolumn{2}{c|}{Publication}                                                                                                                      & CVPR'23               & ICCV'21               & Ours       \\ 
\midrule
Category                                                                & Covariate                                                                   & GaitBase              & PSTA                  & BigGait                         \\ 
\midrule
\multirow{12}{*}{\begin{tabular}[c]{@{}c@{}}Two\\ Mixed\end{tabular}}   & CL-UB                                                                       & $25.2$                  & $73.4$                  & $70.7$                            \\
                                                                        & HVBX-BG                                                                     & $52.1$                  & $94.5$                  & $92.7$                            \\
                                                                        & BG-TC                                                                       & $58.1$                  & $93.3$                  & $92.4$                            \\
                                                                        & SF-CL                                                                       & $36.1$                  & $82.0$                  & $87.0$                            \\
                                                                        & UTR-BX                                                                      & $51.0$                  & $88.9$                  & $90.2$                            \\
                                                                        & DTR-BK                                                                      & $55.1$                  & $93.1$                  & $93.3$                            \\
                                                                        & DTS-HVBX                                                                    & $42.6$                  & $91.9$                  & $92.4$                            \\
                                                                        & UTS-BG                                                                      & $46.8$                  & $90.5$                  & $91.6$                            \\
                                                                        & BM-CL                                                                       & $35.2$                  & $79.6$                  & $86.0$                            \\
                                                                        & CV-HVBX                                                                     & $61.0$                  & $94.3$                  & $93.3$                            \\
                                                                        & CL-CF                                                                       & $39.2$                  & $88.4$                  & $88.6$                            \\ 
\cmidrule{2-5}
                                                                        & \textit{Average}                                                            & $45.7$                  & $88.2$                  & $\mathbf{88.9}$                   \\ 
\midrule
\multirow{12}{*}{\begin{tabular}[c]{@{}c@{}}Three\\ Mixed\end{tabular}} & CL-UB-BG                                                                    & $23.4$                  & $71.3$                  & $69.1$                            \\
                                                                        & BX-BG-CL                                                                    & $35.1$                  & $82.9$                  & $84.1$                            \\
                                                                        & BG-TC-CL                                                                    & $34.3$                  & $81.8$                  & $85.0$                            \\
                                                                        & SF-UB-BG                                                                    & $36.4$                  & $83.5$                  & $82.2$                            \\
                                                                        & UTR-HVBX-CL                                                                 & $31.8$                  & $77.0$                  & $83.9$                            \\
                                                                        & DTR-BK-BG                                                                   & $49.2$                  & $92.3$                  & $92.9$                            \\
                                                                        & DTS-HVBX-CL                                                                 & $26.4$                  & $74.9$                  & $83.4$                            \\
                                                                        & UTS-BG-CL                                                                   & $25.1$                  & $76.0$                  & $84.9$                            \\
                                                                        & BM-CL-BG                                                                    & $33.0$                  & $78.4$                  & $85.2$                            \\
                                                                        & CV-BX-BG                                                                    & $58.8$                  & $93.7$                  & $93.1$                            \\
                                                                        & UB-BG-FA                                                                    & $28.0$                  & $83.0$                  & $79.2$                            \\ 
\cmidrule{2-5}
                                                                        & \textit{Average}                                                            & $34.7$                  & $81.3$                  & $\mathbf{83.9}$                   \\ 
\midrule
\multirow{9}{*}{\begin{tabular}[c]{@{}c@{}}Four\\ Mixed\end{tabular}}   & CL-UB-BG-FA                                                                 & $16.2$                  & $67.5$                  & $63.8$                            \\
                                                                        & BM-CL-BG-BX                                                                 & $32.2$                  & $76.3$                  & $83.5$                            \\
                                                                        & BG-TC-CL-CV                                                                 & $38.0$                  & $79.4$                  & $86.4$                            \\
                                                                        & DTR-BK-BG-CL                                                                & $32.2$                  & $78.8$                  & $86.9$                            \\
                                                                        & DTS-BX-CL-BG                                                                & $25.6$                  & $73.2$                  & $82.7$                            \\
                                                                        & SF-UB-BG-CL                                                                 & $20.6$                  & $66.7$                  & $69.3$                            \\
                                                                        & BG-TC-CL-ST                                                                 & $11.7$                  & $66.8$                  & $66.5$                            \\
                                                                        & UTS-UB-BG-CL                                                                & $15.8$                  & $61.6$                  & $70.3$                            \\ 
\cmidrule{2-5}
                                                                        & \textit{Average}                                                            & $24.0$                  & $71.3$                  & $\mathbf{76.2}$                   \\ 
\midrule
\multirow{4}{*}{\begin{tabular}[c]{@{}c@{}}Five\\ Mixed\end{tabular}}   & \multirow{2}{*}{\begin{tabular}[c]{@{}c@{}}BG-TC-CL-\\ CV-UB\end{tabular}}  & \multirow{2}{*}{$34.1$} & \multirow{2}{*}{$69.6$} & \multirow{2}{*}{$\mathbf{73.8}$}  \\
                                                                        &                                                                             &                       &                       &                                 \\ 
\cmidrule{2-5}
                                                                        & \multirow{2}{*}{\begin{tabular}[c]{@{}c@{}}UTR-BG-CL-\\ BX-CV\end{tabular}} & \multirow{2}{*}{$31.3$} & \multirow{2}{*}{$73.8$} & \multirow{2}{*}{$\mathbf{84.8}$}  \\
                                                                        &                                                                             &                       &                       &                                 \\
\bottomrule[2pt]
\end{tabular}
}
\label{tab: CCGR Easy Mixed}
\end{table}

\noindent \textbf{Activation Maps}. 
Fig.~\ref{fig: sup_1} obtained by the Grad-CAM~\cite{selvaraju2017grad} algorithm exhibits a supplement for Fig.~6 and shows more activation maps on BigGait and video-based ReID methods~\cite{gu2020appearance,wang2021pyramid,zang2022multidirection}.
The visualization insights are the same as in Fig.~6.

\subsection{Related Issues in Rebuttal}

\noindent \textbf{Q1: BigGait's Representation can be noisy.}
We argue a representation should be validated by the performance statistics drawn from a large test set, instead of one or two visual examples.  
Traditional gait representations could be noisy too, \textit{e.g.}, silhouette includes clothing shapes and/or segmentation errors on in-the-wild imagery.
BigGait does demonstrate superiority in cross-clothing and cross-domain tasks, where background/accessory are different between training vs testing data.
More discussions are in Sec.~4.2.

\noindent \textbf{Q2: Background info inflates BigGait's performance.}
We believe BigGait's superiority is not due to the background, from $3$ observations.  
1) Compared Tab.~5~(a) with (d), including backgrounds in BigGait harms its performance.
2) In Tab.~3 and 4, BigGait outperforms ReID methods by large margins though the latter can see full backgrounds.
3) The visualization of activation maps in Fig.~6 further reflects that BigGait focuses on foreground regions. 

\noindent \textbf{Q3: How to handle color noises?}
BigGait regards texture noises as high-frequency signals.
Thus we assume that here color noises refer to low-frequency ones. 
Thanks to the fitting power of neural networks, training models with cross-clothing pairs can partially learn the immunity to these noises (texture and color). 
But the red box of Fig.~6 shows that it is the high-frequency that still heavily impacts existing RGB-based methods.
Hence, we consider high-frequency textures as the primary challenge, and meanwhile, look forward to further improvements brought by color-specific designs.
Thanks for providing this insight. 
The revision will discuss it in detail.

\noindent \textbf{Q4: Why not directly using $f_{de}$?}
Some gait-related features like the body shape may be partially damaged by the geometrical constraints of the denoising branch, while carefully preserved by the appearance branch.
In light of this, we choose to fuse $f_{de}$ and $f_{ap}$ by attention mechanisms, as shown in Fig.~3 and supported by Tab.~5.


\newpage

{
    \small
    \bibliographystyle{ieeenat_fullname}
    \bibliography{main}
}


\end{document}